\documentclass{article}

 \usepackage[main, final]{neurips_2026}
\usepackage[utf8]{inputenc} % allow utf-8 input
\usepackage[T1]{fontenc}    % use 8-bit T1 fonts
\usepackage[hidelinks]{hyperref}
\usepackage{url}            % simple URL typesetting
\usepackage{booktabs}       % professional-quality tables
\usepackage{amsfonts}       % blackboard math symbols
\usepackage{nicefrac}       % compact symbols for 1/2, etc.
\usepackage{microtype}      % microtypography
\usepackage{xcolor}         % colors

\usepackage{amsmath}
\usepackage{graphicx}
\usepackage{booktabs}
\usepackage{algorithm}
\usepackage{algpseudocode}
\usepackage{multicol}
\usepackage{multirow}
\usepackage{makecell}
\usepackage{pifont} % xmark
\usepackage{colortbl} % rowcolor
\usepackage{float}
\usepackage{stfloats}
\usepackage{wrapfig}
\usepackage{cleveref}

\title{Mix-Opt: Mixed Optimization for Memory-Efficient Personalization of Text-to-Image Diffusion Models}

\newcommand\blfootnote[1]{%
  \begingroup
  \renewcommand\thefootnote{}\footnote{#1}%
  \addtocounter{footnote}{-1}%
  \endgroup
}

\author{Seokeon Choi$^\text{*}$ \quad Sunghyun Park$^\text{*}$ \quad Hyoungwoo Park \quad Jeongho Kim \quad Sungrack Yun \vspace{2mm}\\
Qualcomm AI Research$^\dagger$\\{\texttt{\small\{seokchoi, sunpar, hwoopark, jeonghok, sungrack\}@qti.qualcomm.com}}}

\begin{document}

\maketitle

\blfootnote{$^*$ These authors contributed equally.}
\blfootnote{$\dagger$ Qualcomm AI Research is an initiative of Qualcomm Technologies, Inc.}

\begin{abstract}

Memory-efficient personalization is essential for adapting text-to-image diffusion models while preserving user privacy and operating within the limited computational resources of edge devices.
To this end, we propose Mix-Opt, a novel mixed optimization framework that adaptively chooses between backpropagation on low-resolution images (BP-low) and zeroth-order optimization on high-resolution images (ZO-high), guided by the characteristics of the diffusion process.
As observed in our experiments, BP-low efficiently adapts the model to target-specific features but suffers from structural distortions due to resolution mismatch.
Conversely, ZO-high refines high-resolution details with minimal memory overhead but faces slow convergence when applied without prior adaptation.
By complementing both methods, our framework leverages BP-low for effective personalization while using ZO-high to maintain structural consistency, achieving memory-efficient and high-quality fine-tuning.
To maximize the efficacy of both BP-low and ZO-high, we introduce a timestep-aware probabilistic function that dynamically selects the appropriate optimization strategy during training.
This function mitigates overfitting from BP-low at high timesteps, where structural information is critical, while ensuring that ZO-high is applied more effectively as training progresses.
Experimental results across both U-Net- and DiT-based diffusion models demonstrate that Mix-Opt achieves competitive personalization performance while significantly reducing memory consumption.
Mix-Opt enables scalable, high-quality personalization without modifying the inference process or increasing inference latency.

\end{abstract}
\section{Introduction}
\label{sec:introduction}

\begin{figure}[t!]
    \centering
    \includegraphics[width=1.0\linewidth]{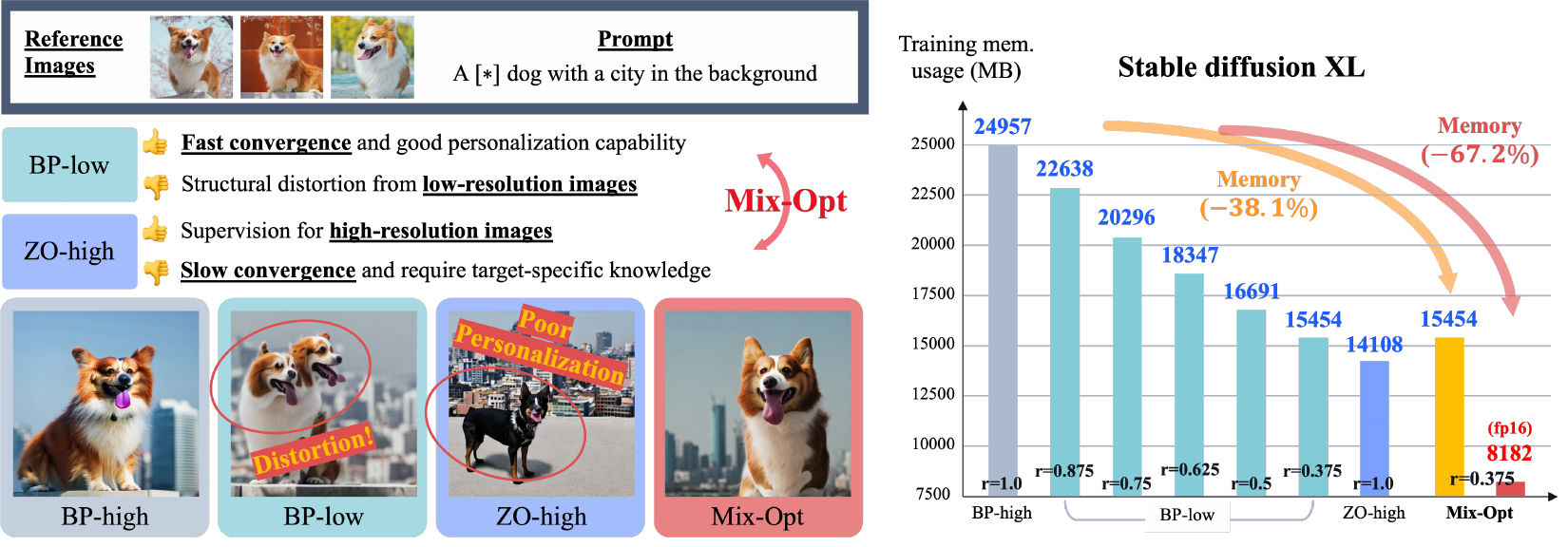}
    \vspace{-0.65cm}
    \caption{Overview of Mix-Opt.
    While BP-low and ZO-high offer memory efficiency, each has its limitations.
    Mix-Opt dynamically combines both methods to achieve high-quality personalization while significantly reducing training memory consumption compared to backpropagation on high-resolution inputs (BP-high).}
    \label{fig:concept}
    \vspace{-0.55cm}
\end{figure}

The increasing popularity of text-to-image diffusion models has driven demand for personalized content generation, enabling applications such as creating avatars~\cite{zeng2023avatarbooth, jiang2024emojidiff} and customizing favorite objects in novel contexts~\cite{chae2023instructbooth}.
To meet this demand, training-free personalization methods~\cite{wei2023elite, li2023blip, shi2024instantbooth, zeng2024jedi} have emerged as efficient solutions, eliminating the need for explicit fine-tuning for each novel concept.
These methods bypass test-time fine-tuning by streamlining the personalization process during inference.
However, they often incur high inference latency, memory overhead, or heavy pre-training requirements, limiting practical deployment under resource constraints.

In contrast, fine-tuning-based methods such as DreamBooth~\cite{ruiz2023dreambooth}, Textual Inversion~\cite{gal2022image}, and Custom Diffusion~\cite{kumari2023multi} update model weights using user-provided images, achieving high-fidelity personalization with fine-grained details.
Due to their high computational demands, these methods are typically executed on servers.
However, growing privacy concerns have increased interest in on-device personalization, which allows users to fine-tune models locally without transmitting sensitive data.
While this approach enhances privacy, the fine-tuning process remains computationally and memory-intensive, posing challenges for deployment on resource-constrained devices.

To address this issue, parameter-efficient fine-tuning techniques such as LoRA~\cite{hu2022lora} have been explored, reducing the number of trainable parameters through low-rank adaptation.
Although effective, LoRA still relies on full backpropagation, which requires storing activations and gradients and therefore incurs significant memory overhead.
To further reduce this overhead, the Memory-Efficient Zeroth-Order Optimizer (MeZO)~\cite{malladi2023fine} has been explored.
MeZO eliminates the need for backpropagation by estimating gradients using only forward passes, substantially reducing memory consumption and making it suitable for resource-constrained personalization.

\begin{figure}[t!]
    \centering
    \includegraphics[width=1.0\linewidth]{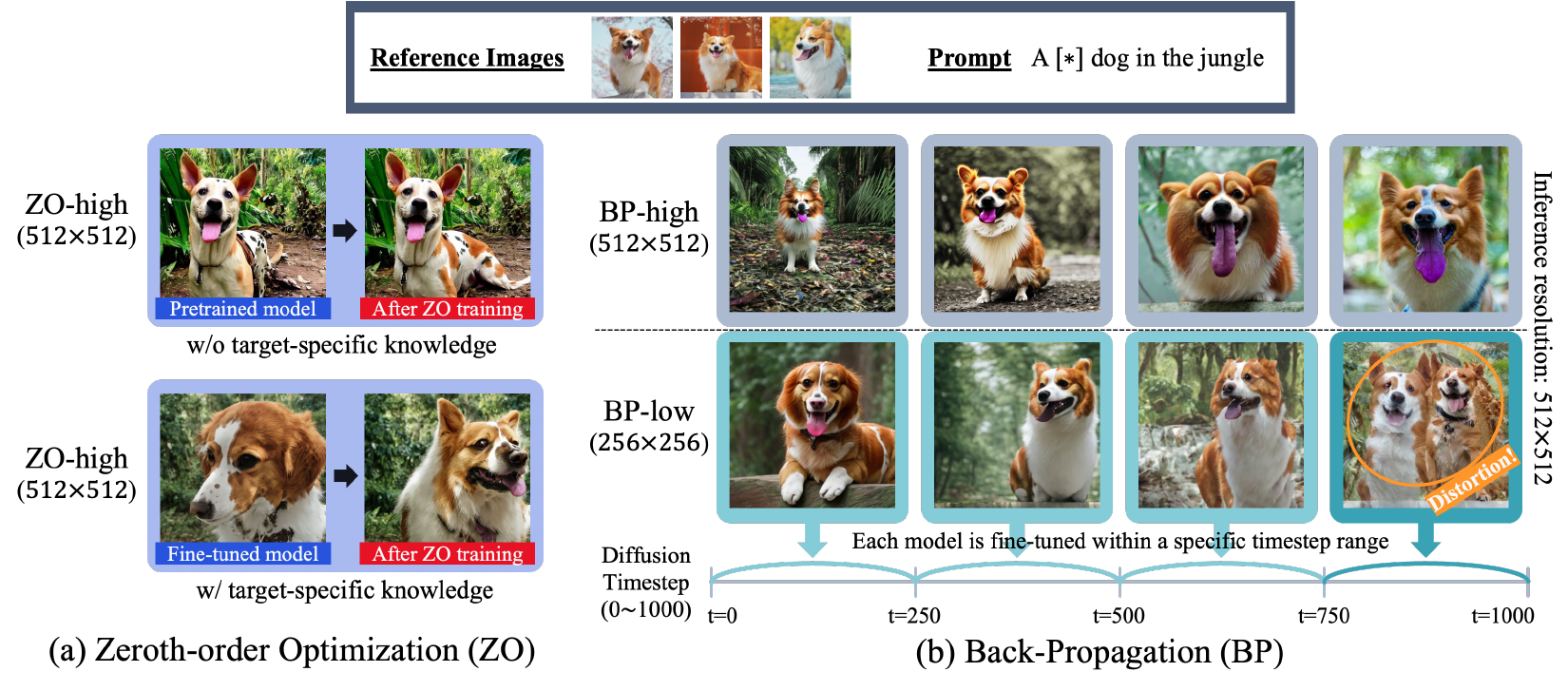}
    \vspace{-0.6cm}
    \caption{Observations for both optimization strategies. (a) MeZO requires target-specific knowledge; applying it from scratch fails to personalize the model, while using it after partial fine-tuning improves personalization. (b) Fine-tuning results for BP-high and BP-low trained at different timestep ranges. BP-high retains personalization features across all timesteps, whereas BP-low struggles to generate consistent outputs due to resolution mismatch, with severe structural distortions at high timesteps.}
    \vspace{-0.4cm}
   \label{fig:observation}
\end{figure}

However, applying MeZO directly to a pretrained model leads to inaccurate gradient estimates and slow convergence due to the absence of target-specific cues.
Similar limitations were reported in ZOODiP~\cite{seo2025efficient}, which required over 30{,}000 iterations for personalization using zeroth-order optimization alone.
Importantly, we also observe that MeZO becomes more effective when preceded by partial fine-tuning that introduces target-specific information (\textbf{Observation 1}), as shown in Fig.~\ref{fig:observation} (a).
High-resolution backpropagation (BP-high) would be ideal for introducing such information, but it is highly memory-intensive.
A practical alternative is to use backpropagation on low-resolution images (BP-low), which reduces memory consumption by decreasing activation sizes.
However, BP-low often produces distorted outputs during generation due to the mismatch between the training and inference resolutions (\textbf{Observation 2}).
This issue is particularly severe when training at higher timesteps (750--1000), where diffusion models primarily recover global structures~\cite{choi2022perception}.
Repeated optimization in this range can bias the model toward low-resolution structural statistics, making BP-low gradients ineffective for maintaining global consistency at the default inference resolution.
T-LoRA~\cite{soboleva2025t} also reported structural overfitting at higher timesteps even with full-resolution inputs, supporting our interpretation that BP-low can be particularly risky in this timestep range.

To overcome these limitations, we propose Mix-Opt, a novel memory-efficient framework that combines two complementary strategies: 1) backpropagation on low-resolution images (BP-low) for lightweight target-specific adaptation, and 2) zeroth-order optimization on high-resolution images (ZO-high) for structural refinement, as illustrated in Fig.~\ref{fig:concept}.
Based on our observations, BP-low effectively introduces target-specific features, whereas ZO-high maintains structural details using high-resolution inputs.
However, BP-low can distort global structures at higher timesteps, while ZO-high has limited impact during the early stage of training because of inaccurate gradient estimates.
To address these trade-offs, we design a timestep-aware probabilistic function that dynamically selects between the two optimization strategies according to the diffusion timestep and training progress.
Specifically, the function favors BP-low at lower timesteps to capture target-specific cues, while increasingly applying ZO-high during the later stages of training to refine structural fidelity.
This approach enables memory-efficient personalization without modifying the sampling process.
We evaluate Mix-Opt on both U-Net- and DiT-based diffusion models, including Stable Diffusion, SANA, and FLUX, to examine its applicability across different backbone architectures.

Our contributions are summarized as follows:
\begin{itemize}
    \item We propose Mix-Opt, a novel mixed optimization framework that combines BP-low and ZO-high for memory-efficient personalization, providing an effective strategy for leveraging zeroth-order optimization in diffusion model personalization.
    \item Based on observations of the complementary behaviors of BP-low and ZO-high, we introduce a timestep-aware probabilistic function that dynamically selects the optimization strategy according to the diffusion timestep and training progress.
    \item We validate Mix-Opt across both U-Net- and DiT-based diffusion models, including Stable Diffusion, SANA, and FLUX.
    Mix-Opt achieves competitive personalization performance while significantly reducing peak optimization memory, including reductions to 2.08 GB on SD V2.1 and 8.18 GB on SDXL, without modifying the inference process or increasing inference latency.
\end{itemize}

\section{Related Work}
\label{sec:related_work}

\noindent\textbf{Text-to-Image Diffusion Models}
% \label{rw:diffusion}
Text-to-image diffusion models have become a cornerstone for generating high-quality images. 
Stable Diffusion~\cite{rombach2022high} introduced diffusion in compressed latent spaces, reducing computational costs while maintaining image quality. It refined the UNet~\cite{ronneberger2015u} backbone and text-conditioning mechanisms, enhancing output fidelity. Recent advancements, such as SDXL~\cite{podell2023sdxl} and SSD-1B~\cite{gupta2024progressive}, expanded model capacity and training data, improving generalization and image quality. More recently, DiT~\cite{peebles2023scalable}-based architectures, such as Stable Diffusion 3~\cite{esser2024scaling}, SANA~\cite{xie2024sana}, and FLUX~\cite{flux2024}, adopted transformer-based backbones, offering improved scalability at the cost of increased memory usage. While these advancements enhance performance, they introduce memory overhead, making efficient fine-tuning essential for deployment on resource-constrained environments. 
% Autoregressive approaches further exacerbate this issue, requiring even larger memory footprints and thus remaining impractical for on-device personalization.

\noindent\textbf{Personalization of Diffusion Models}
% \label{rw:personalization}
Personalization in diffusion models can be categorized into training-based and training-free methods. Training-based approaches, such as Textual Inversion~\cite{gal2022image}, DreamBooth~\cite{ruiz2023dreambooth}, and Custom Diffusion~\cite{kumari2023multi}, fine-tune model weights using user-provided data, enabling high-fidelity image generation. Despite their effectiveness, these methods incur high memory and computational costs, limiting deployment on edge devices. In contrast, training-free methods like ELITE~\cite{wei2023elite}, BLIP-Diffusion~\cite{li2023blip}, InstantBooth~\cite{shi2024instantbooth}, and JeDi~\cite{zeng2024jedi}, inject personalization cues at inference time without modifying model weights. While these methods avoid fine-tuning, they often increase inference latency and memory usage. Our approach enhances training-based methods by providing a memory-efficient fine-tuning strategy that minimizes resource consumption while maintaining personalization fidelity.

% MLLM기반의 방법들은 메모리가 너무 커서 고려대상이 아니라는것도 넣어야하나

\noindent\textbf{Memory-Efficient Fine-Tuning}
% \label{rw:mem_efficient}
Reducing memory consumption during fine-tuning is critical for enabling on-device personalization of large-scale diffusion models. LoRA~\cite{hu2022lora} reduces trainable parameters by focusing on low-rank subspaces of model weights, but still relies on full backpropagation, leading to significant memory overhead. HollowedNet~\cite{cho2025hollowed} further optimizes memory usage by removing LoRA modules from non-essential UNet layers, but remains coupled to backpropagation and the UNet architecture. Memory-efficient Zeroth-order Optimizer (MeZO)~\cite{malladi2023fine} eliminates backpropagation by estimating gradients through forward passes, reducing memory consumption to inference-level requirements. While promising for large language models, its direct application to diffusion models struggles with personalization quality due to the lack of target-specific knowledge. ZOODiP~\cite{seo2025efficient} further illustrated this limitation, requiring over 30,000 iterations for personalization using zeroth-order optimization alone. Building on these foundations, we introduce Mix-Opt, a hybrid optimization framework that leverages the memory efficiency of MeZO while incorporating backpropagation at low resolution (BP-low) for efficient personalization, representing a meaningful advance toward effective ZO utilization.

\section{Proposed Method}
\label{sec:method}

We propose Mix-Opt, a memory-efficient optimization framework for personalization of text-to-image diffusion models.
It combines low-resolution backpropagation (BP-low) and high-resolution zeroth-order optimization (ZO-high) using a timestep-aware selection strategy.
This hybrid approach enables high-quality fine-tuning with low memory usage.

\subsection{Overview of the Optimization Framework}

Given images $\mathcal{X}$ and a prompt $c$ (\textit{e.g.}, ``a scs dog''), the goal is to fine-tune a pretrained diffusion model $\theta$ to generate images that reflect user-specific content.
At each training step $i$, an image $x$ is sampled from $\mathcal{X}$, and a diffusion timestep $t$ is drawn from a uniform distribution $t \sim \mathcal{U}(0, T)$.
Based on $t$ and $i$, a probabilistic function selects either BP-low or ZO-high. Importantly, Mix-Opt does not alter the standard timestep sampling; it only uses $(i,t)$ to choose the optimization mode.

BP-low is used at lower timesteps to efficiently inject target-specific features, while ZO-high helps preserve global structure and reduce overfitting, especially in the later phase of training.
These design choices are motivated by two empirical observations: (1) MeZO is more effective when preceded by partial fine-tuning that introduces target-specific information, and (2) BP-low tends to distort global structure at high timesteps due to resolution mismatch.

\subsection{Backpropagation on Low-Resolution Images}

Backpropagation is applied to low-resolution images $x_{\text{low}}$ (\textit{e.g.}, $256 \times 256$) to reduce memory usage while capturing coarse personalization features.
The optimization follows the standard gradient descent update:
\begin{equation}
    \theta_{i+1} = \theta_i - \eta \nabla_\theta \mathcal{L}_{\text{bp}}(\theta_i; x_{\text{low}}, c, t),
\end{equation}
\noindent
where $\theta_i$ denotes the parameters at training step $i$, $\eta$ is the learning rate, and $\mathcal{L}_{\text{bp}}$ is the diffusion reconstruction loss at timestep $t$.

\begin{algorithm}[t!]
\caption{Training Algorithm}
\label{alg:mixopt}
\begin{algorithmic}[1]
\Require Pretrained diffusion model $\theta_0$, personalization images $\mathcal{X}$, prompt $c$, total number of training steps $i_{\text{max}}$, maximum timestep $t_{\text{max}}$, number of perturbations $N$
\For{training step $i = 1$ to $i_{\text{max}}$}
    \State Sample image $x \in \mathcal{X}$
    \State Sample diffusion timestep $t \sim \mathcal{U}(0, t_{\text{max}})$
    \State Compute $p^{zo}_{i, t}$ using Eq.~\ref{eq:probability_it}
    \If{random() $> p^{zo}_{i, t}$}  \Comment{BP-low}
        \State Downsample $x$ to $x_{\text{low}}$
        \State Compute gradient $\nabla_\theta \mathcal{L}_{\text{bp}}(\theta_i; x_{\text{low}}, c, t)$
        \State Update $\theta_{i+1} = \theta_i - \eta \nabla_\theta \mathcal{L}_{\text{bp}}$
    \Else \Comment{ZO-high}
        \State Use high-resolution image $x$
        \State Initialize accumulated gradient $\hat{g}_i = 0$
        \For{$n = 1$ to $N$}
            \State Sample perturbation $\boldsymbol{z}^{(n)}$
            \State Compute $\hat{g}_i^{(n)}$ using Eq.~\ref{eq:ZO-high}
            \State Accumulate $\hat{g}_i = \hat{g}_i + w_n \hat{g}_i^{(n)}$
        \EndFor
        \State Update $\theta_{i+1} = \theta_i - \alpha \hat{g}_i$
    \EndIf
\EndFor
\end{algorithmic}
\end{algorithm}

\subsection{MeZO on High-Resolution Images}

MeZO~\cite{malladi2023fine} estimates gradients using only forward passes by perturbing parameters and contrasting the resulting losses, following the Simultaneous Perturbation Stochastic Approximation (SPSA)~\cite{spall2002multivariate}.
To improve stability and reduce noise in gradient estimation, we adopt an n-SPSA strategy that averages multiple perturbations, enabling more reliable updates without increasing memory overhead.

For each original image $x$ (\textit{e.g.}, $512 \times 512$), MeZO computes the accumulated gradient.
For each perturbation $n \in \{1, 2, \dots, N\}$, we sample a random perturbation vector $\boldsymbol{z}^{(n)} \sim \mathcal{N}(0, I_d)$ from a standard normal distribution and compute the gradient estimate:
\begin{equation} \label{eq:ZO-high}
    \hat{g}_i^{(n)} = \frac{\mathcal{L}(\theta_i + \epsilon \boldsymbol{z}^{(n)}; x, c, t) - \mathcal{L}(\theta_i - \epsilon \boldsymbol{z}^{(n)}; x, c, t)}{2\epsilon} \boldsymbol{z}^{(n)}.
\end{equation}
The final gradient estimate $\hat{g}_i$ is obtained by a weighted sum of the $N$ estimates, $\hat{g}_i = \sum_{n=1}^{N} w_n \hat{g}_i^{(n)}$, where $w_n = \frac{1}{N}$.
The parameter update is given by $\theta_{i+1} = \theta_i - \alpha \hat{g}_i$, with $\alpha$ as the learning rate for MeZO updates.

\begin{wrapfigure}{r}{0.52\linewidth}
  \centering
  \vspace{-0.3cm}
  \includegraphics[width=\linewidth]{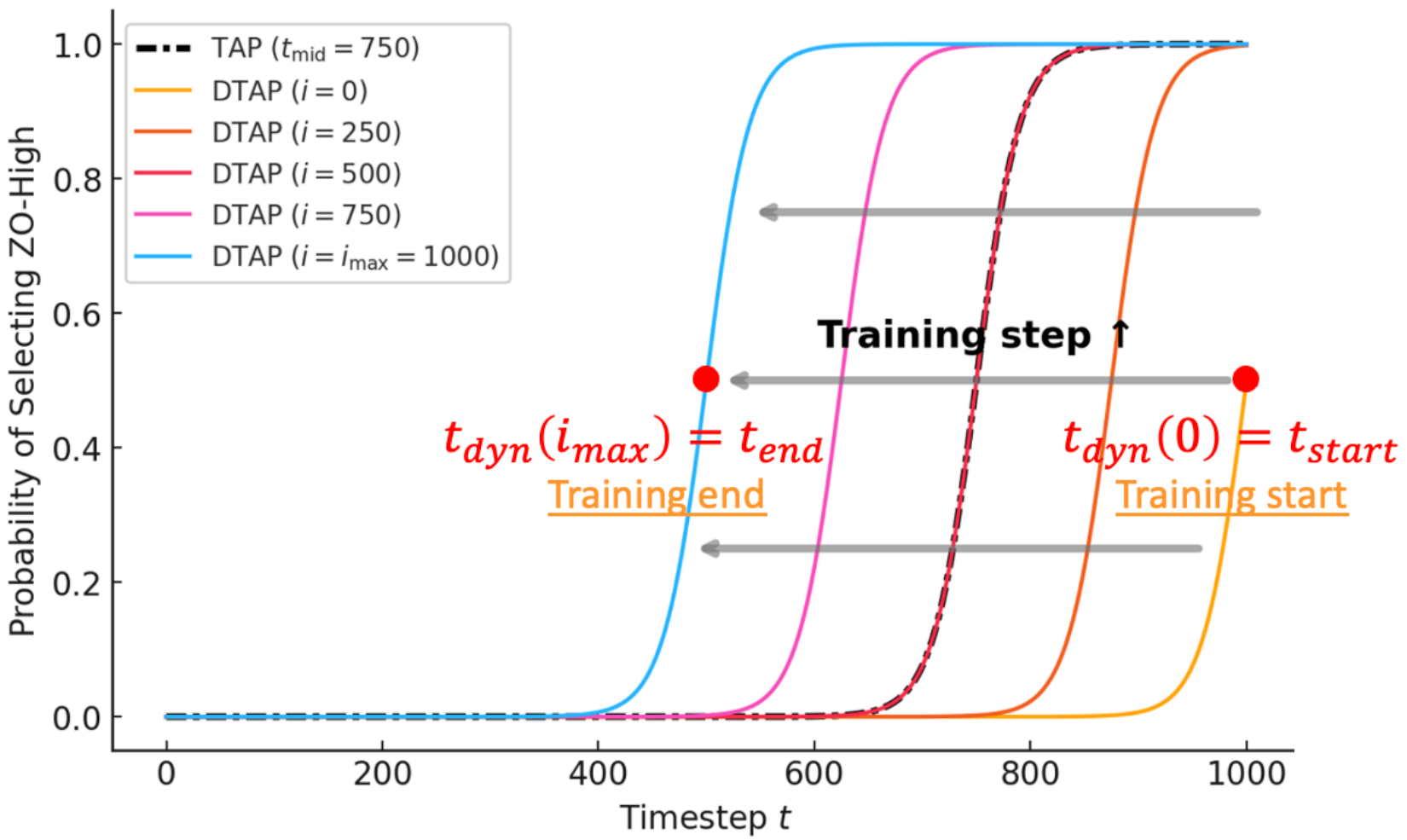}
  \vspace{-0.5cm}
  \caption{Comparison of TAP and DTAP over training steps ($t_{\text{max}}, i_{\text{max}}=1000, k=0.05, t_{\text{mid}}=750$). The black dashed line shows TAP with a fixed $t_{\text{mid}}$, while colored curves show DTAP adapting over training. At $i=i_{\text{mid}}$, DTAP coincides with TAP.}
  \label{fig:probabilistic_function}
  \vspace{-0.7cm}
\end{wrapfigure}

\subsection{Timestep-Aware Probabilistic Function}

% A key component of our framework is the probabilistic function that selects BP-low or ZO-high at each training step. This function adapts the optimization strategy based on the diffusion timestep $t$ and training step $i$, helping to reduce overfitting and structural artifacts.
To reduce overfitting and structural artifacts, our probabilistic function selects the optimization strategy (BP-low or ZO-high) at each step according to the diffusion timestep $t$ and training step $i$.

\subsubsection{Timestep-Aware Probability}

Motivated by \textbf{Observation 2}, BP-low tends to distort global structure at high timesteps ($t \approx t_{\text{max}}$), where structural information is critical~\cite{choi2022perception}.
Accordingly, we define the timestep-aware probability (TAP) of selecting ZO-high as:
\begin{equation}\label{eq:probability_t}
    p^{zo}_t = \frac{1}{1 + e^{-k(t - t_{\text{mid}})}},
\end{equation}\noindent
where $k > 0$ controls the steepness, and $t_{\text{mid}}$ is the midpoint of the transition.
This ensures ZO-high is more likely to be selected at higher timesteps, while BP-low becomes less frequent.

\subsubsection{Dynamic Timestep-Aware Probability}

To reflect \textbf{Observation 1} that MeZO becomes more effective as training progresses, we make the midpoint dynamic in the training step $i$, allowing $t_{\text{mid}}$ to shift from $t_{\text{start}}$ to $t_{\text{end}}$:
\begin{equation}\label{eq:t_dyn}
    t_{\text{dyn}}(i) = t_{\text{start}} + \frac{i}{i_{\text{max}}} (t_{\text{end}} - t_{\text{start}}).
\end{equation}
\noindent
To ensure dynamic timestep-aware probability (DTAP) aligns with TAP such that $t_{\text{dyn}}(i_{\text{mid}}) = t_{\text{mid}}$, we control the transition using $t_{\text{mid}}$ by setting $t_{\text{start}} = t_{\text{max}}$ and $t_{\text{end}} = 2t_{\text{mid}} - t_{\text{max}}$. The final probability becomes:
\begin{equation}\label{eq:probability_it}
     p^{zo}_{i, t} = \frac{1}{1 + e^{-k (t - t_{\text{dyn}}(i))}} .
\end{equation}
During the early stage of training, DTAP favors BP-low over a broad range of timesteps so that the model can efficiently acquire target-specific features.
As training progresses, the transition boundary shifts toward lower timesteps, increasing the frequency of ZO-high updates and improving structural consistency at the default resolution.
% This adaptive selection leverages BP-low in early iterations for fast, low-memory personalization, and gradually shifts toward ZO-high to reduce overfitting and maintain structural integrity, resulting in high-quality personalization with strong memory efficiency. 
% For detailed analysis and selection results based on this function, please refer to \textit{Supplementary Section B}.

\section{Experiments}
\label{sec:experiments}

\subsection{Experimental Setup}  
\label{sec:experiment_setup}  

\noindent\textbf{Implementation Details:}
We implemented Mix-Opt on top of the DreamBooth~\cite{ruiz2023dreambooth} LoRA module using Stable Diffusion V2.1 (SD V2.1)~\cite{rombach2022high}, which served as the primary backbone for most experiments, covering ablation studies, user studies, and hyperparameter tuning (with ratio $r=0.5$).
To evaluate the scalability of Mix-Opt, we extended our analysis to models suitable for edge-device adaptation, including Stable Diffusion V1.5 (SD V1.5)~\cite{rombach2022high}, SDXL~\cite{podell2023sdxl}, SSD-1B~\cite{gupta2024progressive}, SANA~\cite{xie2024sana}, and FLUX~\cite{flux2024}.
All experiments were conducted on NVIDIA A100 GPUs.
For each training sample, we precomputed low- and default-resolution image latents for BP-low and ZO-high, respectively, together with the corresponding text embeddings.
The frozen VAE and text encoder were then offloaded from GPU memory, and the cached representations were reused throughout optimization.
For all experiments, we set the total number of training steps to $i_{\text{max}}=1000$, the LoRA rank to $R=4$, the maximum diffusion timestep to $t_{\text{max}}=1000$, and the learning rate to $\eta=1 \times 10^{-4}$.
The batch size was set to 1, and the resolution matched each model's default setting: $512\times512$ for SD V1.5 and SD V2.1, and $1024\times1024$ for SDXL and SSD-1B.
For Mix-Opt, DTAP was used as the default selection strategy, with key parameters set as follows: $t_{\text{mid}}=750$, $k=0.05$, MeZO learning rate $\alpha=1 \times 10^{-2}$, and gradient accumulation steps $N=4$.

\noindent\textbf{Dataset:}
% All experiments were conducted using the DreamBooth dataset~\cite{ruiz2023dreambooth}, which contains 30 subjects, each paired with 25 unique prompts. Each subject consists of 4 to 6 images, with a total of 9 live subjects and 21 objects. For each subject-prompt pair, four images were generated, and the final scores were computed by averaging the results across all generations. In total, this setup yields 3,000 generated images, which were used to comprehensively evaluate fine-tuning performance.
All experiments were conducted using the DreamBooth dataset~\cite{ruiz2023dreambooth}, which contains 30 subjects, each paired with 25 unique prompts. Each subject consists of 4 to 6 images, with a total of 9 live subjects and 21 objects. For each subject-prompt pair, four images were generated using different random seeds, and the final scores were computed by averaging the results across all generations. In total, this setup yields 3,000 generated images, providing a comprehensive basis for evaluating fine-tuning performance.

\begin{table}[t!]
    \centering
    \footnotesize
    % --- 첫 번째 테이블 (왼쪽) ---
    \begin{minipage}{0.49\textwidth}
        \centering
        \resizebox{\linewidth}{!}{%
        \begin{tabular}{c||l|ccc|c}
            \toprule 
            & \textbf{Method} & \multicolumn{3}{c|}{\textbf{Metrics}} & \textbf{Mem.} \\ 
            & \textbf{(SD V1.5)} & \textbf{DINO} & \textbf{CLIP-I} & \textbf{CLIP-T} & \textbf{(GB)} \\
            \midrule
            & TI & 0.5590 & 0.7780 & 0.2850 & 6.75 \\
            & GF-TI & 0.0110 & 0.5400 & 0.2530 & 2.37 \\
            \multirow{-3}{*}{\rotatebox{90}{\textbf{TI}}} & ZooDiP & 0.5580 & 0.7720 & 0.2870 & 2.37 \\
            \midrule
            \rowcolor{cyan!10} \cellcolor{white} & Full fine-tuning & 0.7149 & 0.8153 & 0.2887 & 17.32 \\
            & PEQA & 0.6040 & 0.7910 & 0.2750 & 6.31 \\
            & TuneQDM & 0.5550 & 0.7880 & 0.2890 & 8.96 \\
            & LoRA-BP-High & 0.6434 & \textbf{0.7926} & 0.2890 & 5.15 \\
            & LoRA-MeZO & 0.2218 & 0.6046 & 0.2645 & 3.63 \\
            \rowcolor{gray!10} \cellcolor{white} & \textbf{Ours \scriptsize (r=0.750)} & 0.6403 & \underline{0.7921} & \underline{0.2904} & 4.44 \\
            \rowcolor{gray!10} \cellcolor{white} & \textbf{Ours \scriptsize (r=0.500)} & \underline{0.6437} & 0.7885 & \underline{0.2904} & 3.91 \\
            \rowcolor{gray!10} \cellcolor{white} & \textbf{Ours \scriptsize (r=0.375)} & 0.6114 & 0.7796 & 0.2847 & 3.73 \\
            \rowcolor{gray!25} \cellcolor{white} & \textbf{Ours\textsuperscript{\dag} \scriptsize (r=0.750)} & 0.6424 & 0.7918 & \textbf{0.2913} & 2.44 \\
            \rowcolor{gray!25} \cellcolor{white} & \textbf{Ours\textsuperscript{\dag} \scriptsize (r=0.500)} & \textbf{0.6448} & 0.7891 & 0.2893 & \underline{2.07} \\
            \rowcolor{gray!25} \cellcolor{white} \multirow{-11}{*}{\rotatebox{90}{\textbf{DB}}} & \textbf{Ours\textsuperscript{\dag} \scriptsize (r=0.375)} & 0.6059 & 0.7764 & 0.2851 & \textbf{1.94} \\
            \bottomrule
        \end{tabular}
        }
        \caption{Comparison of personalization methods on SD V1.5. $\dag$ denotes results obtained with mixed precision (fp16).}
        \label{tab:model_comparison_sd1}
    \end{minipage}
    \hfill
    % --- 두 번째 테이블 (오른쪽) ---
    \begin{minipage}{0.49\textwidth}
        \centering
        \resizebox{\linewidth}{!}{%
        \begin{tabular}{c||l|ccc|c}
            \toprule 
            & \textbf{Method} & \multicolumn{3}{c|}{\textbf{Metrics}} & \textbf{Mem.} \\ 
            & \textbf{(SD V2.1)} & \textbf{DINO} & \textbf{CLIP-I} & \textbf{CLIP-T} & \textbf{(GB)} \\
            \midrule
            \rowcolor{cyan!10} \cellcolor{white} & Full fine-tuning & 0.6889 & 0.7989 & 0.3039 & 17.47 \\
            & HollowedNet \scriptsize (H-30) & 0.6690 & 0.8050 & \underline{0.3030} & 4.18 \\
            & HollowedNet \scriptsize (H-39) & 0.6600 & 0.8050 & 0.3000 & 3.88 \\
            & HollowedNet \scriptsize (H-57) & 0.6180 & 0.7820 & 0.3020 & 3.16 \\
            & LoRA-BP-High & 0.6762 & \textbf{0.8075} & 0.2961 & 5.09 \\
            & LoRA-MeZO & 0.3688 & 0.6664 & \textbf{0.3217} & 3.66 \\
            \rowcolor{gray!10} \cellcolor{white} & \textbf{Ours \scriptsize (r=0.750)} & 0.6758 & 0.8058 & 0.2969 & 4.42 \\
            \rowcolor{gray!10} \cellcolor{white} & \textbf{Ours \scriptsize (r=0.500)} & \textbf{0.6833} & 0.8067 & 0.2938 & 3.92 \\
            \rowcolor{gray!10} \cellcolor{white} & \textbf{Ours \scriptsize (r=0.375)} & 0.6532 & 0.7893 & 0.2950 & 3.75 \\
            \rowcolor{gray!25} \cellcolor{white} & \textbf{Ours\textsuperscript{\dag} \scriptsize (r=0.750)} & \underline{0.6789} & 0.8065 & 0.2965 & 2.44 \\
            \rowcolor{gray!25} \cellcolor{white} & \textbf{Ours\textsuperscript{\dag} \scriptsize (r=0.500)} & 0.6776 & \underline{0.8069} & 0.2926 & \underline{2.08} \\
            \rowcolor{gray!25} \cellcolor{white} \multirow{-12}{*}{\rotatebox{90}{\textbf{DB}}} & \textbf{Ours\textsuperscript{\dag} \scriptsize (r=0.375)} & 0.6526 & 0.7922 & 0.2947 & \textbf{1.82} \\
            \bottomrule
        \end{tabular}}
        \caption{Comparison of personalization methods on SD V2.1. $\dag$ denotes results obtained with mixed precision (fp16).}
        \label{tab:model_comparison_sd2}
    \end{minipage}
    \vspace{-0.5cm}
\end{table}

\noindent\textbf{Evaluation Metrics:}
For quantitative evaluation, we used CLIP-I, CLIP-T~\cite{clip}, and DINO~\cite{dino} scores. CLIP-I measures similarity to reference subject images, CLIP-T evaluates text-image alignment, and DINO assesses subject consistency. Training memory usage was reported in each experiment to assess resource efficiency. VAE and text encoder components were precomputed and excluded from training, and peak memory was measured using \textit{torch.cuda.max\_memory\_allocated}.

\subsection{Scalability across Diffusion Architectures}

% We evaluate the scalability of Mix-Opt across four representative diffusion models using the DreamBooth dataset. \cref{tab:model_comparison_sd1,tab:model_comparison_sd2,tab:model_comparison_sdxl,tab:model_comparison_ssd1b} report image generation performance and training memory usage.
% , while \cref{tab:lora_variation} analyzes model-agnostic behavior under different DreamBooth configurations.

\noindent\textbf{Stable Diffusion V1.5 in Table~\ref{tab:model_comparison_sd1}:}
TI~\cite{gal2022image}-based methods such as GF-TI~\cite{fei2023gradient} and ZooDiP~\cite{seo2025efficient} show low memory usage but limited personalization fidelity.
In particular, ZooDiP combines quantization and zeroth-order optimization, achieving 2.37 GB memory usage, but its reliance on textual inversion leads to lower CLIP-I and DINO scores.
ZooDiP also requires over 30,000 steps to converge, while Mix-Opt reaches better performance in just 1,000 steps, despite also utilizing zeroth-order optimization. 
Among DreamBooth-based methods, full fine-tuning achieves high performance but requires excessive memory, limiting its applicability on edge devices.
Quantization-based approaches (PEQA~\cite{kim2023memory}, TuneQDM~\cite{ryu2024memory}) reduce memory relative to full fine-tuning, but incur notable fidelity loss and remain less competitive than Mix-Opt in both fidelity and memory efficiency.
Furthermore, MeZO~\cite{malladi2023fine} alone fails to achieve robust personalization in DreamBooth settings, highlighting the limitations of relying solely on zeroth-order optimization.
Mix-Opt addresses this by combining ZO with BP-based updates in a dynamic framework. We report Mix-Opt results for $r \in \{0.75, 0.5, 0.375\}$, and find that $r = 0.5$ offers the best trade-off between fidelity and memory usage in SD V1.5.
When $r = 0.5$, Mix-Opt achieves performance comparable to LoRA-BP-High while using \textbf{23.0\%} less memory.
When mixed precision ($\dag$) with fp16 is applied, Mix-Opt reduces the memory requirement by \textbf{59.8\%} relative to LoRA-BP-High while maintaining comparable personalization performance.

\noindent\textbf{Stable Diffusion V2.1 in Table~\ref{tab:model_comparison_sd2}:}
Compared to SD V1.5, SD V2.1 yields higher overall fidelity scores.
We compare Mix-Opt with HollowedNet~\cite{cho2025hollowed}, which reports three variants (H-30, H-39, H-57) based on pruning extent.
H-57 uses the least memory (3.16 GB) but shows severe performance drop.
Under comparable memory budgets, Mix-Opt consistently outperforms HollowedNet in DINO and CLIP-I scores, demonstrating its robustness without requiring architecture-specific structural pruning.
Unlike HollowedNet, which must be manually redesigned for each backbone, Mix-Opt is model-agnostic and supports flexible resolution control, making it more practical.
Moreover, HollowedNet requires additional precomputing for intermediate activations and doubles the inference steps, whereas Mix-Opt operates without additional precomputing and leaves the inference process entirely unchanged, further highlighting its efficiency and practicality.

\begin{table}[t!]
\centering
\footnotesize
% --- 왼쪽 테이블 (SDXL) ---
\begin{minipage}{0.48\textwidth}
    \centering
    \vspace{-0.2cm}
    \resizebox{\linewidth}{!}{%
    \begin{tabular}{c||l|ccc|c}
        \toprule 
        & \textbf{Method} & \multicolumn{3}{c|}{\textbf{Performance}} & \textbf{Mem.} \\ 
        & \textbf{(Base: SDXL)} & \textbf{DINO} & \textbf{CLIP-I} & \textbf{CLIP-T} & \textbf{(GB)} \\
        \midrule
        & LoRA-BP-High & \textbf{0.7397} & \underline{0.8145} & 0.3005 & 24.96 \\
        & LoRA-MeZO & 0.4349 & 0.6670 & \textbf{0.3261} & 14.11 \\
        \rowcolor{gray!10} \cellcolor{white} & \textbf{Ours \scriptsize (r=0.750)} & 0.7310 & 0.8093 & \underline{0.3012} & 20.30 \\
        \rowcolor{gray!10} \cellcolor{white} & \textbf{Ours \scriptsize (r=0.500)} & \underline{0.7350} & 0.8137 & 0.2948 & 16.69 \\
        \rowcolor{gray!10} \cellcolor{white} & \textbf{Ours \scriptsize (r=0.375)} & 0.7302 & 0.8121 & 0.2913 & 15.45 \\
        \rowcolor{gray!25} \cellcolor{white} & \textbf{Ours\textsuperscript{\dag} \scriptsize (r=0.750)} & 0.7269 & 0.8084 & 0.3003 & 11.49 \\
        \rowcolor{gray!25} \cellcolor{white} & \textbf{Ours\textsuperscript{\dag} \scriptsize (r=0.500)} & 0.7336 & 0.8135 & 0.2941 & \underline{9.05} \\
        \rowcolor{gray!25} \cellcolor{white} \multirow{-8}{*}{\rotatebox{90}{\textbf{DB}}} & \textbf{Ours\textsuperscript{\dag} \scriptsize (r=0.375)} & 0.7322 & \textbf{0.8146} & 0.2898 & \textbf{8.18} \\
        \bottomrule
    \end{tabular}}
    \caption{Comparison of personalization methods on SDXL. $\dag$ denotes results obtained with mixed precision (fp16).}
    \label{tab:model_comparison_sdxl}
\end{minipage}
\hfill % 두 테이블 사이의 간격을 최대한 벌림
% --- 오른쪽 테이블 (SSD-1B) ---
\begin{minipage}{0.48\textwidth}
    \centering
    \resizebox{\linewidth}{!}{%
    \begin{tabular}{c||l|ccc|c}
        \toprule 
        & \textbf{Method} & \multicolumn{3}{c|}{\textbf{Performance}} & \textbf{Mem.} \\ 
        & \textbf{(Base: SSD-1B)} & \textbf{DINO} & \textbf{CLIP-I} & \textbf{CLIP-T} & \textbf{(GB)} \\
        \midrule
        & LoRA-BP-High & 0.6841 & 0.7971 & \underline{0.3070} & 15.41 \\
        & LoRA-MeZO & 0.3900 & 0.6539 & \textbf{0.3109} & 9.10 \\
        \rowcolor{gray!10} \cellcolor{white} & \textbf{Ours \scriptsize (r=0.750)} & 0.6964 & 0.8027 & 0.3047 & 12.57 \\
        \rowcolor{gray!10} \cellcolor{white} & \textbf{Ours \scriptsize (r=0.500)} & \textbf{0.7145} & \textbf{0.8106} & 0.3001 & 10.48 \\
        \rowcolor{gray!10} \cellcolor{white} & \textbf{Ours \scriptsize (r=0.375)} & 0.7057 & 0.8048 & 0.2994 & 9.75 \\
        \rowcolor{gray!25} \cellcolor{white} & \textbf{Ours\textsuperscript{\dag} \scriptsize (r=0.750)} & 0.6913 & 0.7985 & 0.3048 & 7.14 \\
        \rowcolor{gray!25} \cellcolor{white} & \textbf{Ours\textsuperscript{\dag} \scriptsize (r=0.500)} & \underline{0.7080} & \underline{0.8075} & 0.2996 & \underline{5.67} \\
        \rowcolor{gray!25} \cellcolor{white} \multirow{-8}{*}{\rotatebox{90}{\textbf{DB}}} & \textbf{Ours\textsuperscript{\dag} \scriptsize (r=0.375)} & 0.7041 & 0.8040 & 0.2989 & \textbf{5.14} \\
        \bottomrule
    \end{tabular}}
    \caption{Comparison of personalization methods on SSD-1B. $\dag$ denotes results obtained with mixed precision (fp16).}
    \label{tab:model_comparison_ssd1b}
\end{minipage}
\vspace{-0.5cm}
\end{table}

\begin{table}[t!]
    \centering
    \footnotesize
    \resizebox{0.7\linewidth}{!}{%
    \begin{tabular}{ll|ccc|c}
        \toprule
        \textbf{Backbone} & \textbf{Method} & \textbf{DINO} & \textbf{CLIP-I} & \textbf{CLIP-T} & \textbf{Mem. (GB)} \\
        \midrule
        & LoRA-BP-High & \textbf{0.7239} & \textbf{0.8043} & 0.3057 & 3.97 \\
        \rowcolor{gray!10} \cellcolor{white} \multirow{-2}{*}{SANA-0.6B} & \textbf{Ours} & 0.7076 & 0.7975 & \textbf{0.3076} & \textbf{1.75} \\
        \midrule
        & LoRA-BP-High & \textbf{0.7125} & \textbf{0.7942} & \textbf{0.3103} & 6.22 \\
        \rowcolor{gray!10} \cellcolor{white} \multirow{-2}{*}{SANA-1.6B} & \textbf{Ours} & 0.7039 & 0.7924 & 0.3073 & \textbf{3.43} \\
        \midrule
        & LoRA-BP-High & \textbf{0.6717} & \textbf{0.7991} & 0.2667 & 55.61 \\
        \rowcolor{gray!10} \cellcolor{white} \multirow{-2}{*}{FLUX.1-dev} & \textbf{Ours} $(r=0.5)$ & 0.6716 & 0.7911 & \textbf{0.2922} & \textbf{33.51} \\
        \bottomrule
    \end{tabular}
    }
    \caption{Comparison of LoRA-BP-High and Mix-Opt on DiT-based models. The results are obtained with mixed precision (bf16).}
    \label{tab:model_comparison_dit}
    \vspace{-0.6cm}
\end{table}

\begin{figure*}[t!]
    \centering
    \includegraphics[width=1.0\linewidth]{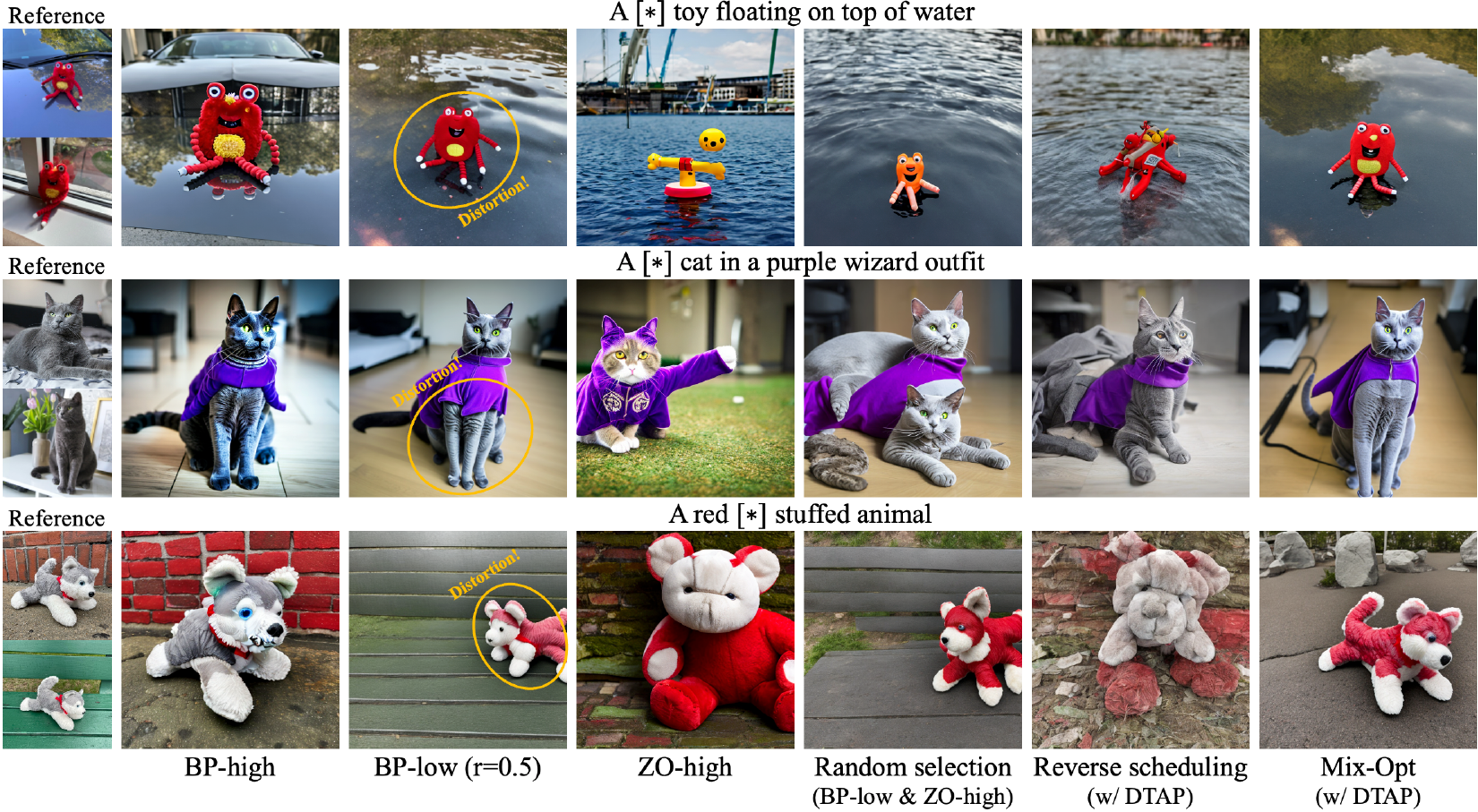}
    \vspace{-0.6cm}
    \caption{Qualitative comparison of different optimization methods, including BP-high, BP-low, ZO-high, Random Selection, Reverse Scheduling, and Mix-Opt (DTAP), conducted with Stable Diffusion v2.1 at low resolution ratio = 0.5.
% BP-high suffers from overfitting, BP-low introduces structural distortions, and ZO-high fails to preserve subject details. Random Selection struggles with instability, causing inconsistent personalization. Additionally, reversing our timestep-aware probability function (setting $k = -0.05$) leads to severe artifacts, confirming the importance of our selection strategy. In contrast, Mix-Opt effectively balances personalization and structure refinement.
}
\vspace{-0.3cm}
\label{fig:qualitative_comparison}
\end{figure*}

\noindent\textbf{SDXL in Table~\ref{tab:model_comparison_sdxl} and SSD-1B in Table~\ref{tab:model_comparison_ssd1b}:} For high-resolution models (1024$\times$1024), Mix-Opt maintains fidelity while significantly reducing memory compared to LoRA-BP-High. The higher resolution allows us to safely lower the resize ratio to $r = 0.375$, enabling further memory savings. With mixed precision, training requires only \textbf{8.18 GB} on SDXL and \textbf{5.14 GB} on SSD-1B. This demonstrates that Mix-Opt not only scales to large backbones but also supports high-resolution personalization.

\noindent\textbf{DiT-based Models in Table~\ref{tab:model_comparison_dit}:}
To investigate whether Mix-Opt can be applied beyond U-Net-based diffusion models, we evaluate it on SANA~\cite{xie2024sana} and FLUX~\cite{flux2024}, which represent two different diffusion transformer designs.
SANA uses a no positional encoding (NoPE) design with local spatial processing in its Mix-FFN, while FLUX uses a rotary positional encoding (RoPE).
Mix-Opt reduces training memory by \textbf{55.9\%} on SANA-0.6B and \textbf{44.9\%} on SANA-1.6B while retaining competitive DINO, CLIP-I, and CLIP-T scores.
On FLUX, it reduces memory by \textbf{39.7\%}, with nearly unchanged DINO, lower CLIP-I, and improved CLIP-T relative to LoRA-BP-High.
Together, these results show that Mix-Opt extends beyond U-Net and NoPE-based DiT architectures to the evaluated PE-based DiT model, supporting its applicability across structurally different diffusion backbones.

\subsection{Qualitative and User Study Analysis}

\begin{wrapfigure}{r}{0.55\linewidth}
    \centering
    \vspace{-0.3cm}
    \includegraphics[width=1.0\linewidth]{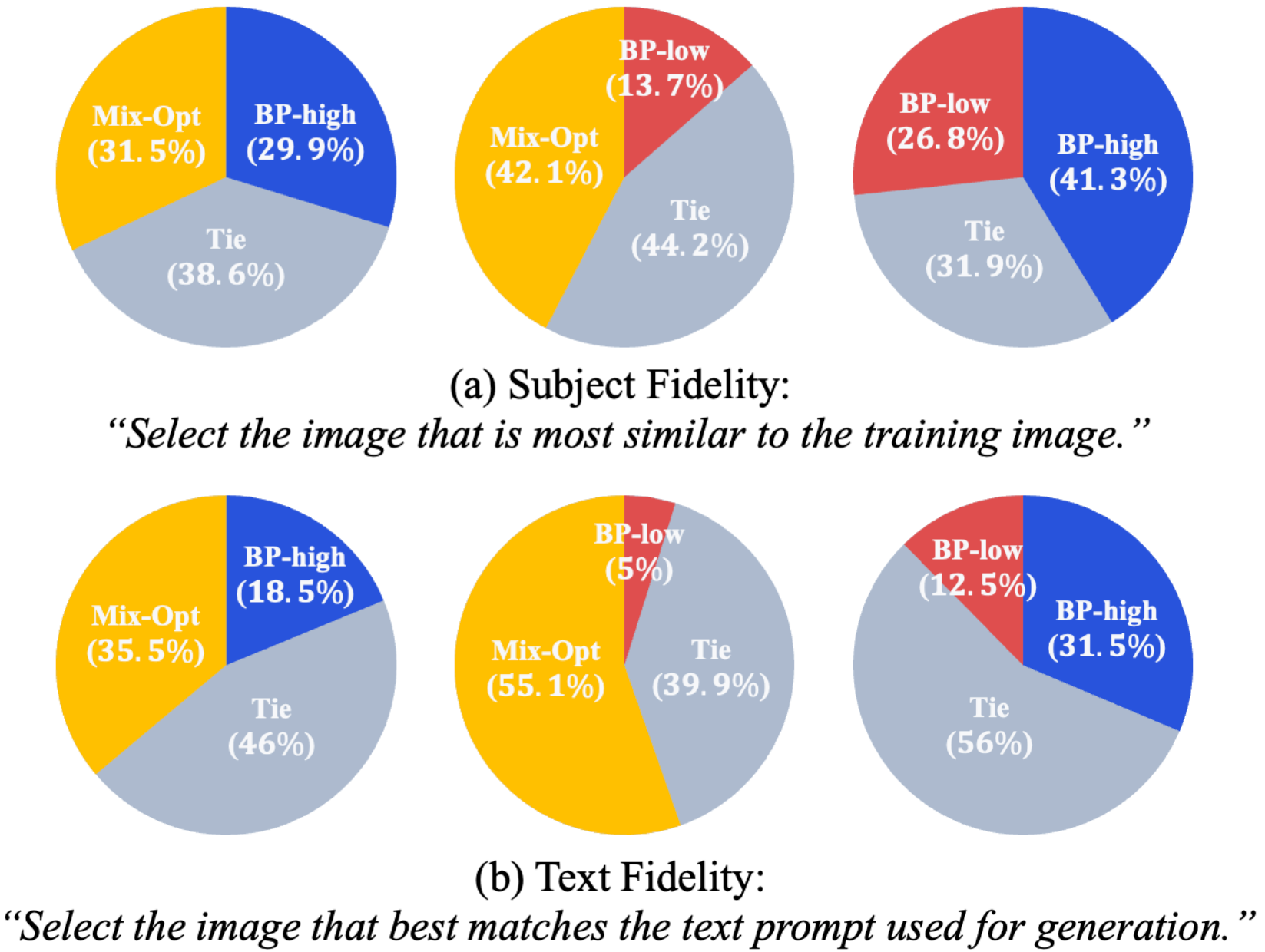}
    \vspace{-0.5cm}
    \caption{User study results on SD V2.1, comparing Mix-Opt with BP-high and BP-low in terms of Subject Fidelity and Text Fidelity. In BP-Low and Mix-Opt, the low-resolution ratio is set to 0.5.}
% Mix-Opt outperforms both BP-high and BP-low, demonstrating superior subject preservation and text alignment.
\vspace{-0.2cm}
\label{fig:user_study}
\end{wrapfigure}
Qualitative comparisons against various optimization strategies are presented in Fig.~\ref{fig:qualitative_comparison}. 
\textbf{BP-high} occasionally suffers from overfitting and saturation-like artifacts, hindering text alignment. 
\textbf{BP-low} exhibits structural distortions, particularly at high timesteps, leading to object deformations and inconsistent subject representation. 
Conversely, \textbf{ZO-high} suffers from slow convergence, failing to learn target-specific features. 
\textbf{Random Selection} lacks a stable optimization path, resulting in suboptimal subject preservation and frequent structural distortions.
% To further evaluate Mix-Opt, we first conducted qualitative comparisons against diverse optimization strategies, as shown in \cref{fig:qualitative_comparison}. \textbf{BP-high} occasionally suffers from overfitting or saturation-like artifacts, which can hinder accurate text alignment. 
% \textbf{BP-low} struggles with structural distortions, particularly at high timesteps, leading to deformations in generated objects and inconsistencies in subject representation. \textbf{ZO-high}, on the other hand, suffers from slow convergence, making it ineffective at learning target-specific features. 
% The \textbf{Random Selection} approach between BP-low and ZO-high lacks a stable optimization strategy, resulting in suboptimal subject preservation and frequent structural distortions. 
\textbf{Reverse Scheduling} refers to a variant where we reversed the selection probability of BP-low and ZO-high ($k = -0.05$). This modification directly contradicts our key observations by prioritizing ZO-high in the early stages and BP-low at high timesteps. This approach resulted in severe structural distortions and failed personalization, confirming that MeZO is ineffective when applied too early and that BP-low exacerbates artifacts when applied at high timesteps. In contrast, \textbf{Mix-Opt} successfully balances BP-low and ZO-high through our DTAP, ensuring structural consistency while preserving fine-grained subject details. In some cases, Mix-Opt even surpasses BP-high in text alignment, demonstrating its ability to alleviate overfitting issues in back propagation.

\begin{table}[t!]
\centering
\footnotesize
% --- 왼쪽 테이블 (Selection Ablation) ---
\begin{minipage}[t]{0.48\textwidth}
    \centering
    \resizebox{\linewidth}{!}{%
    \begin{tabular}{l|ccc}
        \toprule 
        \textbf{Method} & \textbf{DINO} & \textbf{CLIP-I} & \textbf{CLIP-T} \\
        \midrule % Basic
        BP-low & 0.6816 & 0.8099 & 0.2893 \\
        ZO-high & 0.3688 & 0.6664 & 0.3217 \\
        Random selection & 0.6530 & 0.7844 & 0.3050 \\
        \midrule
        Ours (w/ TAP) & 0.6681 & 0.7992 & 0.2948 \\
        Ours (w/ DTAP) & 0.6833 & 0.8067 & 0.2938 \\
        \bottomrule
    \end{tabular}}
    \caption{Ablation study on selection strategies conducted using SD V2.1, with a low-resolution ratio fixed at 0.5.}
    \label{tab:selection_ablation}
\end{minipage}
\hfill % 두 테이블 사이 간격 확보
% --- 오른쪽 테이블 (Impact of t_mid) ---
\begin{minipage}[t]{0.48\textwidth}
    \centering
    \resizebox{\linewidth}{!}{%
    \begin{tabular}{c||l|ccc}
        \toprule 
        & \textbf{Method} & \textbf{DINO} & \textbf{CLIP-I} & \textbf{CLIP-T} \\
        \midrule % TAP
        & $t_{mid} = 250$ & 0.5151 & 0.7210 & 0.3165 \\
        & $t_{mid} = 500$ & 0.6267 & 0.7726 & 0.3060 \\
        \multirow{-3}{*}{\rotatebox{90}{\textbf{TAP}}} 
        & $t_{mid} = 750$ & 0.6681 & 0.7992 & 0.2948 \\
        \midrule % DTAP
        & $t_{mid} = 250$ & 0.5811 & 0.7551 & 0.3112 \\
        & $t_{mid} = 500$ & 0.6327 & 0.7789 & 0.3038 \\
        \multirow{-3}{*}{\rotatebox{90}{\textbf{DTAP}}} 
        & $t_{mid} = 750$ & 0.6833 & 0.8067 & 0.2938 \\
        \bottomrule
    \end{tabular}}
    \caption{Impact of $t_{\text{mid}}$ in TAP and DTAP conducted using SD V2.1 with a low-resolution ratio fixed at 0.5.}
    \label{tab:timestep_ablation}
\end{minipage}
\vspace{-0.7cm}
\end{table}

These findings are validated by a user study with 24 participants (Fig.~\ref{fig:user_study}) using A/B comparisons across all DreamBooth subjects. Mix-Opt consistently outperforms both BP-high and BP-low in subject fidelity and text alignment. Compared to BP-high, Mix-Opt improves text alignment and mitigates instance-level overfitting while maintaining fidelity. Against BP-low, it shows significant gains in both metrics, addressing inherent structural distortions. Notably, BP-low's poor performance relative to BP-high underscores the risks of relying solely on low-resolution optimization for memory efficiency.

\subsection{Ablation Study}

\noindent\textbf{Ablation Study on Selection Strategies:} 
To evaluate the impact of our DTAP, we compared different selection strategies in Table~\ref{tab:selection_ablation}. Interestingly, BP-low achieves competitive DINO and CLIP-I scores but suffers from structural distortions that are not fully captured by these metrics. ZO-high fails to maintain subject fidelity, highlighting its limitations in personalization when applied without adaptation. Random Selection performs between BP-low and ZO-high, underscoring its instability. In contrast, Mix-Opt (w/ \& w/o DTAP) consistently outperforms Random Selection, demonstrating the effectiveness of our DTAP.

We further analyze the effect of $t_{\text{mid}}$ in Table~\ref{tab:timestep_ablation}. Lower $t_{\text{mid}}$ values increase the frequency of MeZO updates while reducing the role of BP, leading to weaker personalization, as indicated by lower CLIP-I and DINO scores. This confirms the importance of BP in establishing target-specific adaptations. Additionally, DTAP consistently surpasses TAP across all $t_{\text{mid}}$ values, demonstrating that dynamically increasing MeZO in later training stages achieves a better balance between personalization and text alignment. Notably, DTAP at $t_{\text{mid}} = 750$ provides the best trade-off, ensuring structural consistency while mitigating overfitting effects.

\begin{figure}[t!]
    \centering
    \includegraphics[width=0.85\linewidth]{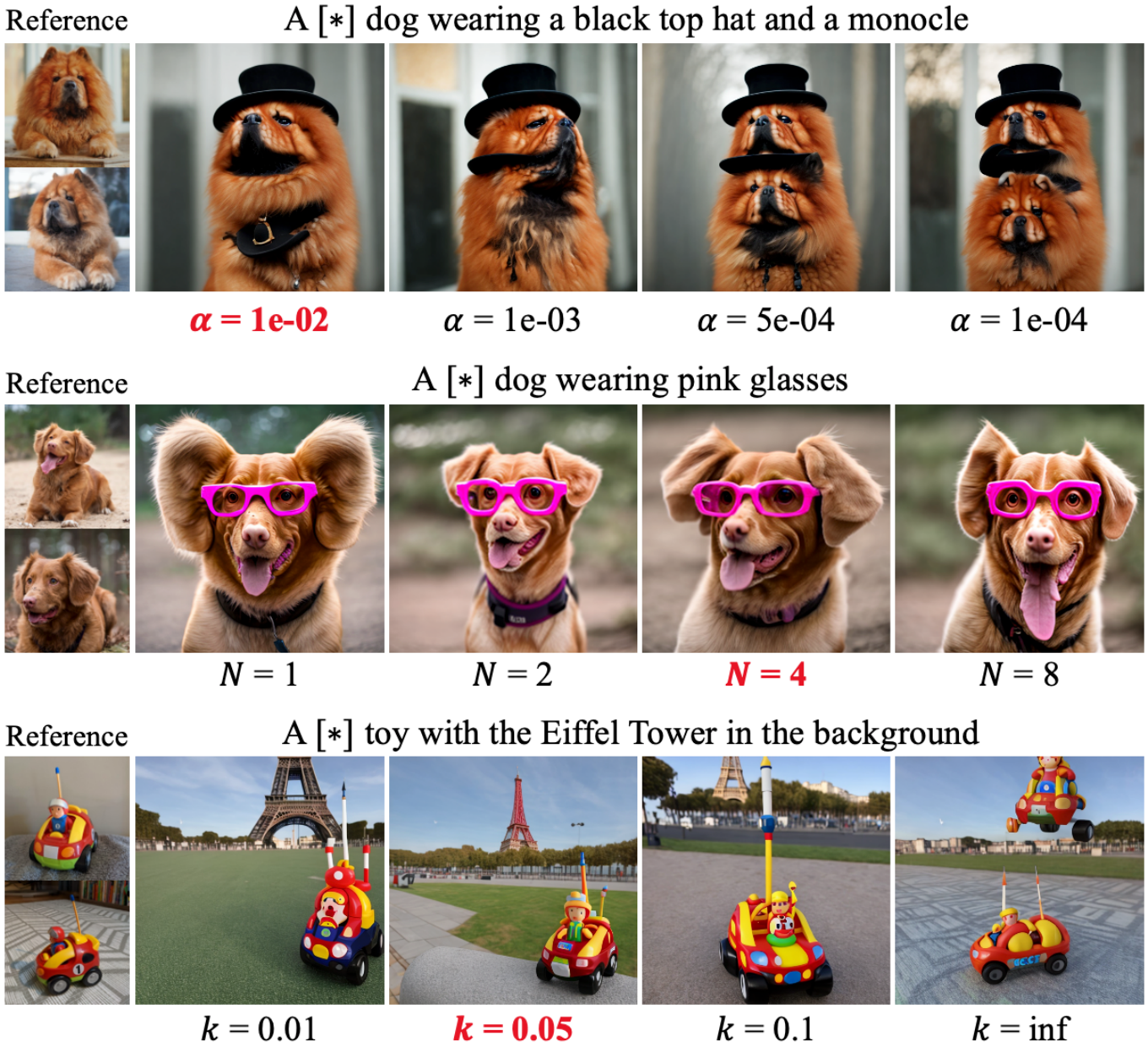}
    \vspace{-0.2cm}
    \caption{Hyperparameter selection for Mix-Opt in SD V2.1, with the low-resolution ratio set to 0.5. We analyze the effects of different choices for the MeZO learning rate $\alpha$, gradient accumulation steps $N$, and the steepness parameter $k$.}
    \vspace{-0.2cm}
    \label{fig:hyperparam_analysis}
\end{figure}

\noindent\textbf{Hyperparameter Selection for Mix-Opt:} 
We conducted an ablation study on Mix-Opt hyperparameters: the learning rate of MeZO ($\alpha$), the gradient accumulation steps ($N$), and the steepness of the DTAP ($k$). The first row of Fig.~\ref{fig:hyperparam_analysis} examines the impact of $\alpha$. A low learning rate prevents ZO-high from effectively refining structural distortions in BP-low, while excessively high values cause optimization instability, leading to gradient exploding or under-personalization. We select $\alpha = 1 \times 10^{-2}$ as the optimal balance. The second row analyzes the effect of $N$. When $N = 1$, the less stable gradient estimation may sometimes lead to slightly reduced personalization performance. Increasing $N$ improves detail refinement, but excessive accumulation significantly slows training without clear performance gains. We find $N = 4$ to provide the best trade-off between stability and efficiency. The third row evaluates $k$, which controls the transition sharpness between BP-low and ZO-high. A high $k$ disrupts structural consistency by introducing abrupt optimization shifts, while a low $k$ flattens the slope of the function, increasing the probability of selecting MeZO too early in training and leading to insufficient personalization. We determine that $k = 0.05$ provides the most balanced optimization strategy, ensuring a smooth transition while maintaining stability.
Although Mix-Opt shows robustness to hyperparameter variations overall, the examples in Fig.~\ref{fig:hyperparam_analysis} highlight extreme cases to illustrate the underlying trends more clearly. In practice, performance remains relatively stable across a reasonable range of values, which makes Mix-Opt a scalable and practical optimization framework for diverse personalization tasks. 
% Please refer to \textit{Supplementary Section C} for the impact of varying $N$ on computational complexity.

\noindent\textbf{Compatibility with 4-bit Weight Quantization.}
We further evaluate whether Mix-Opt can be combined with weight quantization on SD V2.1.
Following QLoRA~\cite{dettmers2023qlora}, we quantize the frozen base-model weights to 4-bit precision using \texttt{bitsandbytes}, while retaining the trainable LoRA parameters, activations, and gradients in fp16.
As shown in Table~\ref{tab:four_bit_quantization}, 4-bit quantization reduces the memory usage of LoRA-BP-High from 5.09 GB to 2.58 GB.
When combined with Mix-Opt, the memory requirement is further reduced to 1.67 GB while maintaining competitive personalization performance.
These results show that Mix-Opt is complementary to base-weight quantization.

\begin{table}[t!]
    \centering
    \footnotesize
    \resizebox{0.7\linewidth}{!}{%
    \begin{tabular}{l|ccc|c}
        \toprule
        \textbf{Method} & \textbf{DINO} & \textbf{CLIP-I} & \textbf{CLIP-T} & \textbf{Mem. (GB)} \\
        \midrule
        LoRA-BP-High & 0.6762 & \textbf{0.8075} & 0.2961 & 5.09 \\
        LoRA-BP-High\textsuperscript{\dag} + 4-bit & 0.6575 & 0.7991 & \textbf{0.2964} & 2.58 \\
        \midrule
        \textbf{Ours\textsuperscript{\dag}} $(r=0.5)$ & \textbf{0.6776} & 0.8069 & 0.2926 & 2.08 \\
        \textbf{Ours\textsuperscript{\dag}} $(r=0.5)$ + 4-bit & 0.6742 & 0.7998 & 0.2922 & \textbf{1.67} \\
        \bottomrule
    \end{tabular}
    }
    \caption{Results with 4-bit weight quantization on SD V2.1. $\dag$ denotes mixed precision with fp16.}
    \label{tab:four_bit_quantization}
    \vspace{-0.4cm}
\end{table}

\section{Conclusion}
\label{sec:conclusion}

In this work, we introduced Mix-Opt, a memory-efficient framework for personalization of text-to-image diffusion models. By integrating BP-low and ZO-high through a timestep-aware probabilistic function, Mix-Opt balances personalization quality and structural consistency while significantly reducing memory consumption. Extensive experiments across multiple diffusion models demonstrated its scalability, generalization, and efficiency, achieving performance comparable to BP-high with substantially lower memory requirements. 
% Our ablation studies further validated the importance of dynamic optimization selection, showing that DTAP consistently improves text alignment and subject fidelity. 
While our study evaluated memory-efficient training on GPUs, an important future direction is deploying Mix-Opt in real on-device environments, optimizing for hardware constraints, and refining inference efficiency.

% \noindent\textbf{Acknowledgement}
% We would like to thank the Qualcomm AI Research team for their valuable discussions.

{
    \small
    \bibliographystyle{plain}
    \bibliography{reference}
}

\newpage
\appendix
\section{Reproducibility}

We provide implementation details and pseudocode in the main paper to ensure reproducibility. All experiments were repeated four times, and the results were averaged. Experiments were conducted using \texttt{diffusers 0.32.0} on top of the DreamBooth LoRA module.

\section{Theoretical Interpretation of DTAP}
\label{app:dtap_derivation}

We provide a relative-risk interpretation of DTAP by considering the errors of the BP-low and ZO-high gradient estimators with respect to the desired high-resolution gradient $g_H(i,t)$.
We define their risks as
\begin{equation}
    R_B(t) = \mathbb{E} \left[ \left\| \hat{g}_B(i,t)-g_H(i,t) \right\|_2^2 \right],
\end{equation}
and
\begin{equation}
    R_Z(i,t) = \mathbb{E} \left[ \left\| \hat{g}_Z(i,t)-g_H(i,t) \right\|_2^2 \right].
\end{equation}
Here, $R_B(t)$ reflects the bias introduced by optimizing a low-resolution surrogate objective, whereas $R_Z(i,t)$ includes the smoothing bias and perturbation variance of the zeroth-order estimator.
The relative risks of the two estimators vary with the diffusion timestep and training progress.
In particular, BP-low becomes less reliable at high timesteps because low-resolution optimization can reinforce inaccurate global structures, whereas ZO-high becomes more effective after the model has acquired target-specific information through partial adaptation.

Using an entropy-regularized soft selection between the two estimators, the probability of selecting ZO-high can be expressed as
\begin{equation}
\label{eq:optimal_zo_selection}
    p_{\mathrm{ZO}}^*(i,t) = \operatorname{sigmoid} \left( \frac{R_B(t)-R_Z(i,t)}{\tau} \right),
\end{equation}
where $\tau$ controls the smoothness of the selection.
Thus, ZO-high is selected more frequently when the relative risk of BP-low is higher.

Around a point at which the two estimators have similar risks, we locally approximate their relative risk as
\begin{equation}
    R_B(t)-R_Z(i,t) \approx a(t-t_0)+b(i-i_0),
\end{equation}
where $a>0$ reflects the increasing relative risk of BP-low at higher timesteps, and $b>0$ reflects the increasing effectiveness of ZO-high as training progresses.
Substituting this approximation into Eq.~\ref{eq:optimal_zo_selection} yields a sigmoid selection probability with a transition boundary that decreases linearly over the training steps.
This corresponds to the form of DTAP used in Eq.~\ref{eq:probability_it}.

This derivation provides a local relative-risk interpretation of DTAP rather than a global optimality guarantee.
The DTAP parameters, including $k$ and $t_{\mathrm{mid}}$, are empirically selected.

\section{Additional Experiments}

\noindent\textbf{Multi-Concept Personalization with TARA.}
We combine Mix-Opt with TARA~\cite{peng2026tara} to evaluate its applicability to multi-concept personalization.
TARA independently trains a concept-specific LoRA module and composes the trained modules during inference.
We replace BP-high with Mix-Opt during the training of each module while retaining the original TARA objective and inference procedure.
As shown in Table~\ref{tab:tara_multiconcept}, Mix-Opt reduces peak optimization memory while maintaining comparable multi-concept personalization performance.
In particular, the $r=0.5$ setting reduces memory from 2.76 GB to 2.07 GB and achieves a higher CLIP-I score than BP-high.

\begin{table}[h!]
    \centering
    \footnotesize
    \resizebox{0.7\linewidth}{!}{%
    \begin{tabular}{l|ccc|c}
        \toprule
        \textbf{Method} & \textbf{DINO} & \textbf{CLIP-I} & \textbf{CLIP-T} & \textbf{Mem. (GB)} \\
        \midrule
        TARA + BP-high\textsuperscript{\dag} & \textbf{0.2744} & 0.6179 & 0.3165 & 2.76 \\
        TARA + Ours\textsuperscript{\dag} $(r=0.75)$ & 0.2653 & 0.6218 & \textbf{0.3173} & 2.44 \\
        TARA + Ours\textsuperscript{\dag} $(r=0.5)$ & 0.2658 & \textbf{0.6252} & 0.3159 & \textbf{2.07} \\
        \bottomrule
    \end{tabular}
    }
    \caption{Multi-concept personalization results with TARA on SD V1.5.
    $\dag$ denotes mixed precision with fp16.}
    \label{tab:tara_multiconcept}
    \vspace{-0.3cm}
\end{table}

\noindent\textbf{Style Personalization with PairCustomization.}
We further combine Mix-Opt with PairCustomization~\cite{jones2024customizing} to evaluate its applicability to style personalization.
We retain the original PairCustomization objective and replace BP-high with Mix-Opt during the training of each style-LoRA.
For evaluation, we use text prompts from StyleAlign~\cite{stylealign} and generate 800 images for each optimization setting.
We report the DreamSim distance~\cite{dreamsim} between the target-style references and generated images, where a lower value indicates better perceptual style alignment.
As shown in Table~\ref{tab:paircustomization}, Mix-Opt reduces peak optimization memory while maintaining comparable style alignment on both SD V1.5 and SD V2.1.

\begin{table}[h!]
    \centering
    \footnotesize
    \resizebox{0.73\linewidth}{!}{%
    \begin{tabular}{ll|cc}
        \toprule
        \textbf{Base Model}
        & \textbf{Method}
        & \textbf{DreamSim} $\downarrow$
        & \textbf{Mem. (GB)} $\downarrow$ \\
        \midrule
        \multirow{3}{*}{SD V1.5}
        & PairCustom + BP-high\textsuperscript{\dag} & 0.7791 & 2.76 \\
        & PairCustom + Ours\textsuperscript{\dag} $(r=0.75)$ & 0.7816 & 2.44 \\
        & PairCustom + Ours\textsuperscript{\dag} $(r=0.5)$ & \textbf{0.7712} & \textbf{2.07} \\
        \midrule
        \multirow{3}{*}{SD V2.1}
        & PairCustom + BP-high\textsuperscript{\dag} & \textbf{0.7901} & 2.77 \\
        & PairCustom + Ours\textsuperscript{\dag} $(r=0.75)$ & 0.8005 & 2.44 \\
        & PairCustom + Ours\textsuperscript{\dag} $(r=0.5)$ & 0.7991 & \textbf{2.08} \\
        \bottomrule
    \end{tabular}
    }
    \caption{Style-personalization results with PairCustomization. Lower DreamSim indicates better style alignment, and $\dag$ denotes mixed precision with fp16.}
    \label{tab:paircustomization}
    \vspace{-0.3cm}
\end{table}

\noindent\textbf{Ablation Study.}
As shown in Table~\ref{tab:lora_variation}, we evaluate Mix-Opt under four DreamBooth configurations by varying LoRA rank ($R \in \{4, 128\}$) and enabling or disabling prior preservation loss to assess model-agnostic performance. Across all settings, Mix-Opt remains stable and effective. At high rank ($R = 128$), LoRA-BP-High occasionally produces saturation-like artifacts, so the number of iterations was reduced to 500. This demonstrates that Mix-Opt can robustly adapt its optimization strategy even when the iteration count changes, dynamically selecting methods as needed. Moreover, with prior preservation loss enabled, Mix-Opt further enhances text alignment and shows robustness even when richer loss functions are incorporated, while still saving memory.

\begin{table}[h!]
    \centering
    \footnotesize
    \begin{tabular}{c||l|ccc|c}
        \toprule 
        & \multicolumn{1}{c|}{\multirow{2}{*}{\textbf{Method}}} & \multicolumn{3}{c|}{\textbf{Performance Metrics}} & \multicolumn{1}{c}{\textbf{Mem.}} \\ 
        & & \textbf{DINO} & \textbf{CLIP-I} & \textbf{CLIP-T} & \textbf{(GB)} \\
        
        \midrule % Rank-4 
        & LoRA-BP-High & 0.6762 & \textbf{0.8075} & \underline{0.2961} & 5.09 \\
        \rowcolor{gray!10} \cellcolor{white} & Ours \scriptsize (r=0.750) & 0.6758 & 0.8058 & \textbf{0.2969} & 4.42 \\
        \rowcolor{gray!10} \cellcolor{white} & Ours \scriptsize (r=0.625) & \underline{0.6787} & 0.8043 & 0.2955 & 4.14 \\
        \rowcolor{gray!10} \cellcolor{white} 
        \multirow{-4}{*}{\rotatebox{90}{\textbf{R-4}}} 
        & Ours \scriptsize (r=0.500) & \textbf{0.6833} & \underline{0.8067} & 0.2938 & 3.92 \\
        
        \midrule % Rank-4 + pp
        & LoRA-BP-High & \textbf{0.6325} & \textbf{0.7803} & 0.3084 & 6.65 \\
        \rowcolor{gray!10} \cellcolor{white} & Ours \scriptsize (r=0.750) & \underline{0.6313} & \underline{0.7772} & \underline{0.3103} & 5.31 \\
        \rowcolor{gray!10} \cellcolor{white} & Ours \scriptsize (r=0.625) & 0.6280 & 0.7723 & \textbf{0.3109} & \underline{4.45} \\
        \rowcolor{gray!10} \cellcolor{white}
        \multirow{-4}{*}{\rotatebox{90}{\textbf{R-4\tiny(pp)}}}
        & Ours \scriptsize (r=0.500) & 0.6126 & 0.7658 & \underline{0.3103} & \textbf{4.32} \\
        
        \midrule % Rank-128 (500 iteration)
        & LoRA-BP-High & 0.7020 & 0.8167 & \textbf{0.2905} & 5.49 \\
        \rowcolor{gray!10} \cellcolor{white} & Ours \scriptsize (r=0.750) & \underline{0.7095} & \underline{0.8170} & \underline{0.2893} & 4.78 \\
        \rowcolor{gray!10} \cellcolor{white} & Ours \scriptsize (r=0.625) & \textbf{0.7121} & \textbf{0.8201} & 0.2886 & \underline{4.48} \\
        \rowcolor{gray!10} \cellcolor{white}
        \multirow{-4}{*}{\rotatebox{90}{\textbf{R-128}}}
        & Ours \scriptsize (r=0.500) & 0.6913 & 0.8132 & 0.2845 & \textbf{4.25} \\

        \midrule % Rank-128 + pp
        & LoRA-BP-High & 0.6520 & 0.7981 & 0.2855 & 7.13 \\
        \rowcolor{gray!10} \cellcolor{white} & Ours \scriptsize (r=0.750) & \underline{0.6793} & \textbf{0.8006} & 0.2897 & 5.72 \\
        \rowcolor{gray!10} \cellcolor{white} & Ours \scriptsize (r=0.625) & \textbf{0.6811} & 0.7977 & \textbf{0.2918} & \underline{5.15} \\
        \rowcolor{gray!10} \cellcolor{white}
        \multirow{-4}{*}{\rotatebox{90}{\textbf{R-128\tiny(pp)}}} 
        & Ours \scriptsize (r=0.500) & 0.6776 & \underline{0.7998} & \underline{0.2909} & \textbf{4.67} \\
        \bottomrule
    \end{tabular}
    \caption{Impact of LoRA rank (R) and prior preservation loss (pp) on performance and memory consumption in SD V2.1. Rank-128 is trained for only 500 iterations due to artifacts in BP-High.}
    \label{tab:lora_variation}
    \vspace{-0.6cm} % 표 아래쪽 여백 조절
\end{table}

\section{Analysis on DTAP Function}

Dynamic Timestep-Aware Probability (DTAP) introduced in the main paper defines the probability of selecting ZO-high at step $(i,t)$ as:
\[
p^{zo}_{i, t} = \frac{1}{1 + e^{-k (t - t_{\text{dyn}}(i))}},
\]
where $t_{\text{dyn}}(i)$ is modeled as a linear function of $i$:
\[
t_{\text{dyn}}(i) = t_{\text{start}} + \frac{i}{i_{\text{max}}} (t_{\text{end}} - t_{\text{start}}) = t_{\text{max}} + \frac{2i}{i_{\text{max}}} (t_{\text{mid}} - t_{\text{max}}),
\]
where $t_{\text{start}} = t_{\text{max}}$ and $t_{\text{end}} = 2 t_{\text{mid}} - t_{\text{max}}$.

\begin{figure}
    \centering
    \includegraphics[width=\linewidth]{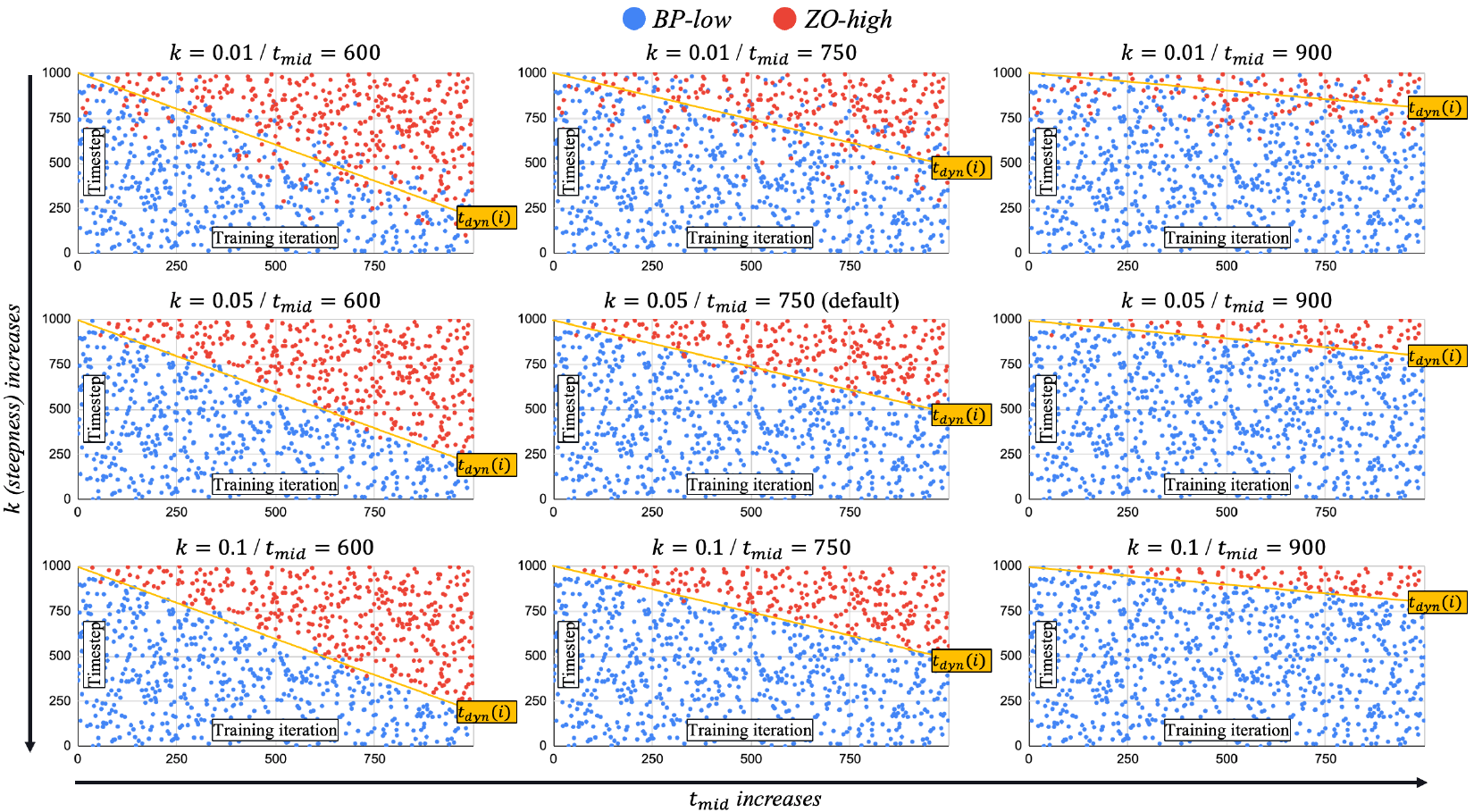}
    \vspace{-0.2cm}
    \caption{DTAP selection patterns across training iterations (x-axis) and timesteps (y-axis). Blue points represent BP-low and red points represent ZO-high. The boundary is defined as a linear function of $i$, given by $t_{\text{dyn}}(i)$, illustrating how $k$ and $t_{\text{mid}}$ influence the transition.}
    \vspace{-0.3cm}
    \label{fig:dtap_selection}
\end{figure}

This design ensures BP-low dominates early iterations while ZO-high becomes more frequent later, mitigating overfitting and structural distortions. Fig.~\ref{fig:dtap_selection} illustrates how DTAP partitions the $(i,t)$ plane under different $(k, t_{\text{mid}})$ settings. The x-axis represents training iteration and the y-axis represents timestep. Blue points indicate BP-low and red points indicate ZO-high. In the experiments, timesteps are uniformly sampled within the range of 0 to 1000. The boundary separating BP and ZO is formed linearly with $i$ according to $t_{\text{dyn}}(i)$, showing how $k$ and $t_{\text{mid}}$ control the transition. Because $t_{\text{start}}$ is set to $t_{\text{max}}$, early iterations behave similarly across settings, but as $t_{\text{mid}}$ decreases, ZO-high is selected more frequently in later training stages. At $i_{\text{mid}}$, the boundary aligns with $t_{\text{mid}}$, confirming the intended scheduling.

\begin{figure*}[t!]
    \centering
    \includegraphics[width=\linewidth]{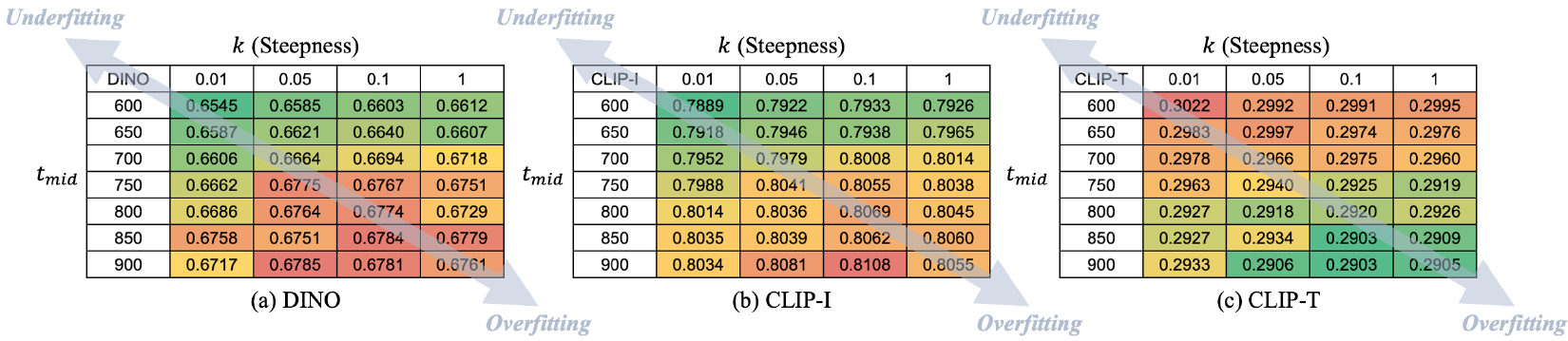}
    \vspace{-0.2cm}
    \caption{Effect of $k$ and $t_{\text{mid}}$ on Mix-Opt performance for SD V2.1 ($r=0.5$). Heatmaps use red for higher scores and green for lower scores. Higher $t_{\text{mid}}$ (lower ZO-high frequency) increases DINO and CLIP-I but reduces CLIP-T, indicating overfitting risks.}
    \vspace{-0.3cm}
    \label{fig:heatmap}
\end{figure*}

We also analyze the frequency of ZO-high selections. Under $k=0.05$ and $i_{\max}=t_{\max}=1000$, the average number of ZO-high selections decreases as $t_{\text{mid}}$ increases: 400 times at $t_{\text{mid}}=600$, 246 times at $t_{\text{mid}}=750$, and 101 times at $t_{\text{mid}}=900$. This reflects the desired scheduling: higher $t_{\text{mid}}$ biases toward BP-low for longer. The parameter $k$ controls the sharpness of the BP-low/ZO-high transition. Larger $k$ yields a sharper boundary and more deterministic selection near $t\approx t_{\text{dyn}}(i)$; smaller $k$ increases stochastic mixing. In the limit, $k\to 0$ approaches random selection, whereas $k\to\infty$ approximates a hard threshold similar to a step function. Both extremes are undesirable; we find $k\in[0.01,0.1]$ robust, with $k=0.05$ used by default.

Fig.~\ref{fig:heatmap} shows performance heatmaps for SD V2.1 with $r=0.5$ across $k$ and $t_{\text{mid}}$. Colors range from red (high score) to green (low score). Higher $t_{\text{mid}}$ (which corresponds to lower ZO-high frequency) tends to increase DINO and CLIP-I but reduce CLIP-T, indicating overfitting risks. Even when object structures are visually distorted (e.g., overlapping shapes), DINO and CLIP-I remain high if local features match the reference, while CLIP-T drops as text alignment deteriorates. These trends confirm that hyperparameter selection based solely on metrics can be misleading; visual inspection is recommended. We typically select $k\in[0.01,0.1]$ and $t_{\text{mid}}\in[700,800]$ for balanced results. When $k=1$, the transition becomes too abrupt, slightly degrading performance as shown in Fig.~\ref{fig:heatmap}. For practical adjustment, increase $(k,t_{\text{mid}})$ when underfitting occurs and decrease them when overfitting or structural distortion appears.

\begin{table}[t!]
    \centering
    \footnotesize
    \begin{tabular}{lcccc}
        \toprule
        \textbf{Training Cost (TFLOPS)} & \textbf{N=1} & \textbf{N=2} & \textbf{N=4} & \textbf{N=8} \\
        \midrule
        Mix-Opt \scriptsize(r=0.750)\normalsize & \cellcolor{gray!20}{1.5026} & \cellcolor{gray!20}{1.8370} & 2.5057 & 3.8432 \\
        Mix-Opt \scriptsize(r=0.625)\normalsize & \cellcolor{gray!20}{1.2417} & \cellcolor{gray!20}{1.5761} & 2.2448 & 3.5823 \\
        Mix-Opt \scriptsize(r=0.500)\normalsize & \cellcolor{gray!20}{1.0283} & \cellcolor{gray!20}{1.3627} & \cellcolor{gray!20}{2.0314} & 3.3689 \\
        Mix-Opt \scriptsize(r=0.375)\normalsize & \cellcolor{gray!20}{0.8623} & \cellcolor{gray!20}{1.1967} & \cellcolor{gray!20}{1.8654} & 3.2029 \\
        \bottomrule
    \end{tabular}
    \caption{Average training computational cost (lower is better) per iteration under Mix-Opt for different $r$ and $N$. Mix-Opt can achieve up to 58\% reduction by lowering $r$ and $N$. BP-high baseline is 2.0622 TFLOPS per iteration. Values lower than BP-high are highlighted.}
    \vspace{-0.2cm}
    \label{tab:tflops}
\end{table}

\begin{table}[t!]
    \centering
    \footnotesize
    \begin{tabular}{lcccc}
    \toprule
    \textbf{Training Time (s)} & \textbf{N=1} & \textbf{N=2} & \textbf{N=4} & \textbf{N=8} \\
        \midrule
        Mix-Opt \scriptsize(r=0.750)\normalsize & \cellcolor{gray!20}{184.5} & \cellcolor{gray!20}{211.9} & 288.1 & 431.2 \\
        Mix-Opt \scriptsize(r=0.625)\normalsize & \cellcolor{gray!20}{189.3} & \cellcolor{gray!20}{231.4} & 294.8 & 431.5 \\
        Mix-Opt \scriptsize(r=0.500)\normalsize & \cellcolor{gray!20}{177.8} & \cellcolor{gray!20}{210.6} & 282.8 & 416.6 \\
        Mix-Opt \scriptsize(r=0.375)\normalsize & \cellcolor{gray!20}{179.1} & \cellcolor{gray!20}{217.7} & 280.1 & 450.1 \\
        \bottomrule
    \end{tabular}
    \caption{Training time (lower is better) over 1000 iterations for different $r$ and $N$. Reducing $N$ provides the largest speedup, while changes in $r$ have a smaller effect. BP-high baseline is 251.4 seconds. Values lower than BP-high are highlighted.}
    \vspace{-0.2cm}
    \label{tab:time}
\end{table}

\begin{table}[t!]
    \centering
    \footnotesize
    \begin{tabular}{lcccc}
    \toprule
    \textbf{Subject Fidelity} & \textbf{N=1} & \textbf{N=2} & \textbf{N=4} & \textbf{N=8} \\
    \midrule
    Mix-Opt \scriptsize(r=0.750)\normalsize & 0.7396 & \cellcolor{gray!20}{0.7434} & 0.7408 & 0.7370 \\
    Mix-Opt \scriptsize(r=0.625)\normalsize & \cellcolor{gray!20}{0.7454} & \cellcolor{gray!20}{0.7420} & 0.7415 & \cellcolor{gray!20}{0.7442} \\
    Mix-Opt \scriptsize(r=0.500)\normalsize & 0.7361 & 0.7355 & \cellcolor{gray!20}{0.7450} & 0.7356 \\
    Mix-Opt \scriptsize(r=0.375)\normalsize & 0.7184 & 0.7171 & 0.7213 & 0.7203 \\
    \bottomrule
    \end{tabular}
    \caption{Subject fidelity (higher is better) measured by the average of DINO and CLIP-I scores across $r$ and $N$. Performance remains stable down to $r=0.5$ with a slight drop at $r=0.375$, and variation in $N$ has minimal impact. BP-high baseline is 0.7419. Values higher than BP-high are highlighted.}
    \vspace{-0.2cm}
    \label{tab:clipi}
\end{table}

\section{Training Computation and Time}

We analyze the training computational cost in terms of average FLOPs per iteration (TFLOPs) and total training time (s). Two key factors influence efficiency: the resolution ratio $r$ for BP-low and the number of MeZO gradient accumulation steps $N$. Table~\ref{tab:tflops} summarizes average TFLOPS per iteration, Table~\ref{tab:time} shows training time over 1000 iterations, and Table~\ref{tab:clipi} reports subject fidelity scores.

Training computation decreases substantially as $r$ and $N$ are reduced, enabling significant efficiency gains compared to BP-high (2.0622 TFLOPS). For example, $(r=0.5, N=1)$ averages 1.0283 TFLOPS (about 50\% reduction), and $(r=0.375, N=1)$ averages 0.8623 TFLOPS (about 58\% reduction). These values are averaged over 1000 iterations under DTAP with $k=0.05$ and $t_{\text{mid}}=750$, where BP-low is selected approximately 754 times and ZO-high 246 times. From the perspective of training time, reducing $N$ provides the largest speedup because MeZO cost scales with the number of forward passes, while changes in $r$ have a smaller effect. Notably, even when $N$ is set to 1 or 2, the training time remains lower than BP-high, highlighting the efficiency advantage of the proposed approach.

As shown in Table~\ref{tab:clipi}, Subject fidelity, measured by average CLIP-I and DINO scores, remains stable down to $r=0.5$ with only a slight drop at $r=0.375$, and variation in $N$ has minimal impact on personalization fidelity. Our default setting $(N=4, r=0.5)$ offers a strong balance between efficiency and performance, though training time under this setting is slightly higher than BP-high. This can be mitigated by lowering $N$ to 2 without noticeable quality loss, providing a practical trade-off for scenarios where training time is critical.

To summarize, across its different variants, Mix-Opt simultaneously reduces training memory, computational cost, and training time compared to BP-high, while maintaining competitive performance. These improvements have direct implications for on-device deployment: smaller training memory enables personalization of larger models, and lower TFLOPS together with shorter training cycles directly translate to reduced energy consumption and thermal load, making personalization practical on resource-constrained hardware. Importantly, Mix-Opt achieves these gains without modifying the inference process, the standard timestep sampling schedule, or the training loss function.

\section{Further Discussions}

We highlight additional discussions and insights gained from our experiments.

\noindent\textbf{Layer-wise ZO Perturbation:}
We explored an alternative ZO-High approach by splitting LoRA layers and independently estimating the approximate gradient for each split. This method aimed to improve gradient estimation by allowing finer-grained updates within the model. However, we observed no significant advantage over applying perturbations across the entire LoRA parameter space. The global perturbation strategy remained the most effective in maintaining stable personalization and refinement while keeping memory consumption low.

\noindent\textbf{Replacing ZO-High with Random Perturbation:}
Instead of using ZO-High to approximate gradients, we tested a purely random perturbation approach. When the perturbation scale was too small, there was little effect on training, while larger perturbations led to gradient exploding. Even when the scale was carefully tuned for fairness in comparison with MeZO, personalization often degraded significantly. In contrast, MeZO effectively incorporates target-specific information, improving personalization performance and serving as a structural refinement mechanism for high-resolution supervision.

\noindent\textbf{Mix-Opt with Full Fine-Tuning:}
We attempted to apply Mix-Opt in a full fine-tuning setting instead of LoRA. While Mix-Opt remained functional, several issues emerged. The backpropagation memory cost of low-resolution updates was significantly higher in full fine-tuning compared to LoRA-based optimization, reducing the effectiveness of our memory-efficient approach. Given that Mix-Opt is primarily designed for resource-efficient training, we focused on LoRA-based fine-tuning, which offers a more practical trade-off between memory consumption and personalization performance.

\noindent\textbf{Downsampling Method:}
We experimented with different downsampling strategies for BP-low, including \textit{resize}, \textit{crop}, and \textit{crop \& resize}. The cropping method often resulted in excessive magnification of the subject, while crop \& resize required significantly more iterations to achieve the same level of personalization as BP-high. Given these trade-offs, we opted for a simple resize operation, which provided a balance between memory efficiency and personalization quality.

\noindent\textbf{Future Directions:}
An important direction is to evaluate Mix-Opt in real on-device environments.
Although our experiments demonstrate substantial reductions in peak optimization memory, practical deployment also requires consideration of hardware-specific constraints, training latency, energy consumption, and numerical precision.
Extending Mix-Opt beyond LoRA to other parameter-efficient fine-tuning techniques may further broaden its applicability.
These directions could improve the practical deployment of memory-efficient personalization across resource-constrained devices and future diffusion architectures.

% \newpage
% \input{checklist.tex}

\end{document}